\pdfoutput=1
\documentclass[11pt]{article}

\usepackage[final]{coling}

\usepackage{times}
\usepackage{latexsym}

\usepackage[T1]{fontenc}
\usepackage[utf8]{inputenc}

\usepackage{microtype}

\usepackage{inconsolata}

\usepackage{graphicx}

\usepackage{amsmath}
\usepackage{amsthm}
\usepackage{tabularx}
\usepackage{booktabs}
\usepackage{float}
\usepackage{graphicx}
\usepackage{amssymb}
\usepackage{bm}
\usepackage{bbm}
\usepackage{xcolor}
\usepackage{textcomp}
\usepackage{mathtools}

\usepackage{algorithm}
\usepackage{algorithmic}
\usepackage{dsfont}
\usepackage{caption}
\usepackage{subcaption}
\usepackage{colortbl}
\usepackage{multirow}
\usepackage{stmaryrd}
\usepackage[normalem]{ulem}

\usepackage{xspace}
\usepackage{comment}
\usepackage{thmtools, thm-restate}
\usepackage{verbatim}

\usepackage{array}
\newcolumntype{R}[1]{>{\raggedleft\arraybackslash}p{#1}}
\newcolumntype{L}[1]{>{\raggedright\arraybackslash}p{#1}}
\newcolumntype{C}[1]{>{\centering\arraybackslash}p{#1}}

\usepackage{amssymb}% http://ctan.org/pkg/amssymb
\usepackage{pifont}% http://ctan.org/pkg/pifont
\newcommand{\cmark}{\textcolor{green}{\ding{51}}}%
\newcommand{\xmark}{\textcolor{red}{\ding{55}}}%

\usepackage[commandnameprefix=always, commentmarkup=uwave, defaultcolor=brown, draft]{changes}
\definechangesauthor[name=You]{EK}
\definechangesauthor[name=Junhao Liu]{JL}
\definechangesauthor[name=Todo, color=blue]{Todo}

\newcommand{\paperacronym}{SOAR} % stands for Single-cell Omics ARena
\newcommand{\benchmarkName}{\paperacronym{}-RNA} % stands for Single-cell Omics ARena
\newcommand{\benchmarkMulName}{\paperacronym{}-MultiOmics} % stands for Single-cell Omics ARena

\newcommand{\llmNumber}{8} % instruction-tuned LLM number
\newcommand{\datasetNumber}{11} % all datasets in both RNA and MultiOmics benchmarks
\newcommand{\taskNumber}{1226} % all question-answering tasks in both RNA and MultiOmics benchmarks

\title{Single-Cell Omics Arena: A Benchmark Study for \\Large Language Models on Cell Type Annotation Using Single-Cell Data}

\author{
 Junhao Liu\textsuperscript{1} \quad Siwei Xu\textsuperscript{1} \quad Lei Zhang\textsuperscript{2} \quad Jing Zhang\textsuperscript{1} 
\\
 \textsuperscript{1}University of California, Irvine \\
 \textsuperscript{2}University of Chinese Academy of Sciences
\\
 \texttt{\{junhao.liu, zhang.jing\}@uci.edu}
}

\begin{document}
\maketitle
\begin{abstract}
Over the past decade, the revolution in single-cell sequencing has enabled the simultaneous molecular profiling of various modalities across thousands of individual cells, allowing scientists to investigate the diverse functions of complex tissues and uncover underlying disease mechanisms. Among all the analytical steps, assigning individual cells to specific types is fundamental for understanding cellular heterogeneity. However, this process is usually labor-intensive and requires extensive expert knowledge. Recent advances in large language models (LLMs) have demonstrated their ability to efficiently process and synthesize vast corpora of text to automatically extract essential biological knowledge, such as marker genes, potentially promoting more efficient and automated cell type annotations. To thoroughly evaluate the capability of modern instruction-tuned LLMs in automating the cell type identification process, we introduce \paperacronym{}, a comprehensive benchmarking study of LLMs for cell type annotation tasks in single-cell genomics. Specifically, we assess the performance of \llmNumber{} instruction-tuned LLMs across \datasetNumber{} datasets, spanning multiple cell types and species. Our study explores the potential of LLMs to accurately classify and annotate cell types in single-cell RNA sequencing (scRNA-seq) data, while extending their application to multiomics data through cross-modality translation. Additionally, we evaluate the effectiveness of chain-of-thought (CoT) prompting techniques in generating detailed biological insights during the annotation process. The results demonstrate that LLMs can provide robust interpretations of single-cell data without requiring additional fine-tuning, advancing the automation of cell type annotation in genomics research.
\end{abstract}

\section{Introduction}

% \textcolor{red}{[JZ2JZ: summary: major contributions. 1 benchmarking different LLM on cell type annotations; 2. introduce COT into this task; 3. cross-modality annotation]}

The recent advancements in single-cell technologies~\cite{stuart2019integrative,ma2020integrative,wu2021babel} enable simultaneous molecular profiling of diverse modalities across tens of thousands of individual cells, allowing researchers to explore the heterogeneity and functionality within complex tissues by uncovering rare or previously unidentified cell types that would otherwise be obscured by traditional bulk tissue sequencing methods. Among the various tasks in single-cell analysis, the classification of cells into known canonical or novel cell types—referred to as cell type annotation—serves as the primary and most fundamental step~\cite{jagadeesh2022identifying}. This is crucial because each cell type performs distinct roles, and accurate identification facilitates the study of their specific contributions to biological processes, development, and disease mechanisms~\cite{eraslan2022single}. However, this task is computationally demanding, labor-intensive, and requires extensive labeling, as traditional methods rely heavily on expert knowledge of gene functions and cell biology to ensure annotation accuracy. Consequently, there is a pressing need to develop efficient and precise cell type annotation methods to automate and streamline this process.

Over the past decade, there have been significant transformations in the acquisition and utilization of domain-specific knowledge required for cell type annotation, largely driven by advancements in artificial intelligence (AI) and natural language processing (NLP)~\cite{levinecell2sentence}. Notably, large language models (LLMs) have emerged as powerful tools for efficiently processing and synthesizing extensive text corpora, including scientific literature, expert discussions, and technical documents, to accurately associate key features—such as marker genes—with specific cell types~\cite{achiam2023gpt,yang2024qwen2, dubey2024llama, deepseek-aiDeepSeekLLMScaling2024,jiang2024mixtral}. This has made automated cell type annotation increasingly feasible with minimal expert involvement. Moreover, LLMs are capable of integrating diverse and complex data types, including genomic datasets, biological knowledge, and previous annotations, to further enhance the accuracy and efficiency of cell type classification~\cite{levinecell2sentence}. Consequently, several recent pioneering studies~\cite{levinecell2sentence, theodoris2023transfer,cui2024scgpt,hou2024assessing} have developed LLM-based automated cell type annotation for single-cell RNA-sequencing (scRNA-seq) data, demonstrating strong concordance with traditional manual annotations~\cite{hu2023cellmarker,aran2019reference,ianevski2022fully}.

Although promising, automating cell type annotation in single-cell sequencing data using diverse modalities via LLMs is still in its nascent stages and faces significant challenges. First, while LLMs have been extensively benchmarked across various domains~\cite{hendrycksmeasuring, cobbe2021training}, there is a notable gap in comprehensive analyses specifically focused on cell type labeling tasks in single-cell omics data. Benchmarking LLMs on existing datasets could provide valuable insights into their performance in key areas such as novel cell type discovery, marker gene identification, cross-condition comparisons, and the optimization of prompt learning strategies. Second, current methods~\cite{levinecell2sentence, theodoris2023transfer,cui2024scgpt} predominantly rely on one-step prompt learning without incorporating intermediate reasoning processes~\cite{kojima2022large}, offering limited understanding of gene function and cell biology during cell type assignment. Finally, unlike scRNA-seq data~\cite{stuart2019integrative}, which measures gene expression levels that are directly interpretable by LLMs, most single-cell sequencing technologies generate domain-specific molecular measurements that are not immediately accessible to these models. For instance, scATAC-seq data~\cite{mimitou2020scalable} profiles open chromatin regions in individual cells, while scHi-C experiments~\cite{lieberman2009comprehensive} output chromatin contact probabilities in three-dimensional space. These specialized features are rarely represented in the textual corpora used to train LLMs, presenting a great challenge for direct cell type annotation on such data types.

To address this gap, we developed \paperacronym{}, the first extensive benchmarking study of LLMs for cell-type annotation tasks using single-cell genomics data. Our comprehensive analysis spans \datasetNumber{} diverse datasets and evaluates \llmNumber{} instruction-tuned LLMs across \taskNumber{} cell-type annotation-related tasks. Additionally, we investigate the effectiveness of chain-of-thought (CoT) prompting techniques to assess their performance in annotation and to extract biological insights from the models using single-cell data with different modalities. The key contributions of our study are summarized follows:
\textbf{(1)} We curated a comprehensive cell type annotation benchmark and evaluation protocol on single cell genomic data, encompassing five species, hundreds of cell types from complex tissues, to thoroughly assess the understanding of single-cell biology using \llmNumber{} popular LLMs. \textbf{(2)} Our results demonstrate that many LLMs exhibit a robust capacity to interpret single-cell RNA-seq data without the need for additional fine-tuning. These models are capable of generating detailed reasoning processes to assist researchers in analyzing single-cell biology data. \textbf{(3)} We further extended the application of LLMs to multiomics data, by employing cross-modality translation models to enhance their analytical capabilities on molecular features that are not immediately interpretable by LLMs.
% \begin{itemize}
%     \item We curated a comprehensive cell type annotation benchmark and evaluation protocol on single cell genomic data, encompassing five species, hundreds of cell types from complex tissues, to thoroughly assess the understanding of single-cell biology using \llmNumber{} popular LLMs. 
%     \item Our results demonstrate that many LLMs exhibit a robust capacity to interpret single-cell RNA-seq data without the need for additional fine-tuning. These models are capable of generating detailed reasoning processes to assist researchers in analyzing single-cell biology data.
%     \item We further extended the application of LLMs to multi-omics data, beyond RNA-seq, by employing cross-modality translation models to enhance their analytical capabilities on molecular features that are not immediately interpret-able by LLMs.
% \end{itemize}

\section{Related Work}

The field of large language models (LLMs) has made substantial progress in recent years. Models such as GPT~\cite{achiam2023gpt, brown2020language}, Qwen~\cite{yang2024qwen2, bai2023qwen}, and the Llama series~\cite{dubey2024llama, touvron2023llama2, touvron2023llama} have demonstrated remarkable performance across a wide range of tasks, including question answering, image captioning, and text summarization. Training an LLM to follow human instructions typically involves two key phases: pre-training on a large text corpus and subsequent instruction tuning~\cite{ouyang2022training}. Once tuned, various evaluation frameworks have been developed to measure the models' adherence to human instructions. For example, MMLU~\cite{hendrycksmeasuring} is widely used to assess a model's general knowledge and reasoning abilities across diverse subjects, while GSM8K~\cite{cobbe2021training} serves as a benchmark for mathematical problem-solving. Despite these advancements, much of the training and evaluation of LLMs continues to focus on natural language processing, limiting their potential to address more complex scientific challenges. 
Single-cell genomics is a groundbreaking technique that enables researchers to quantify molecular features at the level of individual cells, facilitating the study of cellular heterogeneity and functionality within complex tissues. 
However, applying LLMs to automate the cell type labeling process remains challenging, primarily due to the uncertainty surrounding LLMs' ability to accurately interpret the specialized domain knowledge inherent in these datasets.

% The field of LLMs has made significant strides in recent years. Models such as GPT~\cite{achiam2023gpt, brown2020language}, Qwen~\cite{yang2024qwen2, bai2023qwen}, and the LLaMA series~\cite{dubey2024llama, touvron2023llama2, touvron2023llama} have demonstrated impressive capabilities across a broad spectrum of tasks, including question answering, image captioning, and text summarization. Training an LLM to follow human instructions typically involves two key phases: pre-training on a large corpus of text, followed by instruction-tuning~\cite{ouyang2022training}. Once tuned, various evaluation methods have been developed to assess the ability of LLMs to adhere to human instructions. For instance, MMLU~\cite{hendrycksmeasuring} has been widely used to evaluate a model's general knowledge and reasoning abilities across diverse subjects, while GSM8K~\cite{cobbe2021training} serves as a benchmark for mathematical problem-solving tasks. Despite substantial progress, much of the training and evaluation for LLMs are still focusing on natural language processing, limiting their potential for solving advanced scientific problems. Single-cell genomics, for example, is a groundbreaking technique that allows researchers to quantify molecular features in individual cells, enabling the study of cellular heterogeneity and functionality within complex tissues.

Several pioneering works have explored the application of pre-trained models to genomics sequence data, such as Geneformer~\cite{theodoris2023transfer} and scGPT~\cite{cui2024scgpt}. However, these studies focus on training bidirectional contextual genomics embedding models rather than developing general instruction-following large language models capable of reasoning through genomics analysis tasks based on human commands. Consequently, these models require additional post-finetuning to perform specific downstream tasks, aligning with the pretrain-then-finetune paradigm introduced by BERT~\cite{kenton2019bert}, rather than a general instruction-tuned approach. Recently, Cell2Sentence~\cite{levinecell2sentence} introduced a post-pretraining strategy that integrates natural language-based LLMs with transcriptomic knowledge, enabling LLMs to follow human instructions to complete various genomics analysis tasks. However, due to the lack of standardized genomics-related benchmarks, several key questions remain unresolved: (1) What is the ability of contemporary instruction-tuned LLMs to analyze genomic data without additional finetuning? (2) If these LLMs can analyze genomic data, how can their performance in genomics analysis be further enhanced? and (3) Can LLMs handle heterogeneous genomic data beyond text?

To address the aforementioned questions, this work introduces a cell type annotation using single-cell genomics benchmark, \paperacronym{}, to assess the instruction-following capabilities of LLMs in the field of single-cell genomics. Specifically, \paperacronym{} comprises two components: \benchmarkName{} to evaluate different LLMs' performance on scRNA-seq data, and \benchmarkMulName{} focusing on other omic modalities that are not immediately interpretable by LLMs. We thoroughly evaluate the performance of LLMs in solving genomics analysis tasks and propose a zero-shot chain-of-thought prompting strategy, which has been shown to effectively enhance the ability of LLMs to perform genomics analysis. We also introduce a cross-modality translation module to integrate RNA-seq data with other genomic modalities, enabling LLMs to analyze heterogeneous genomic data.
\begin{figure}[t]
    \centering
         \includegraphics[width=0.48\textwidth]{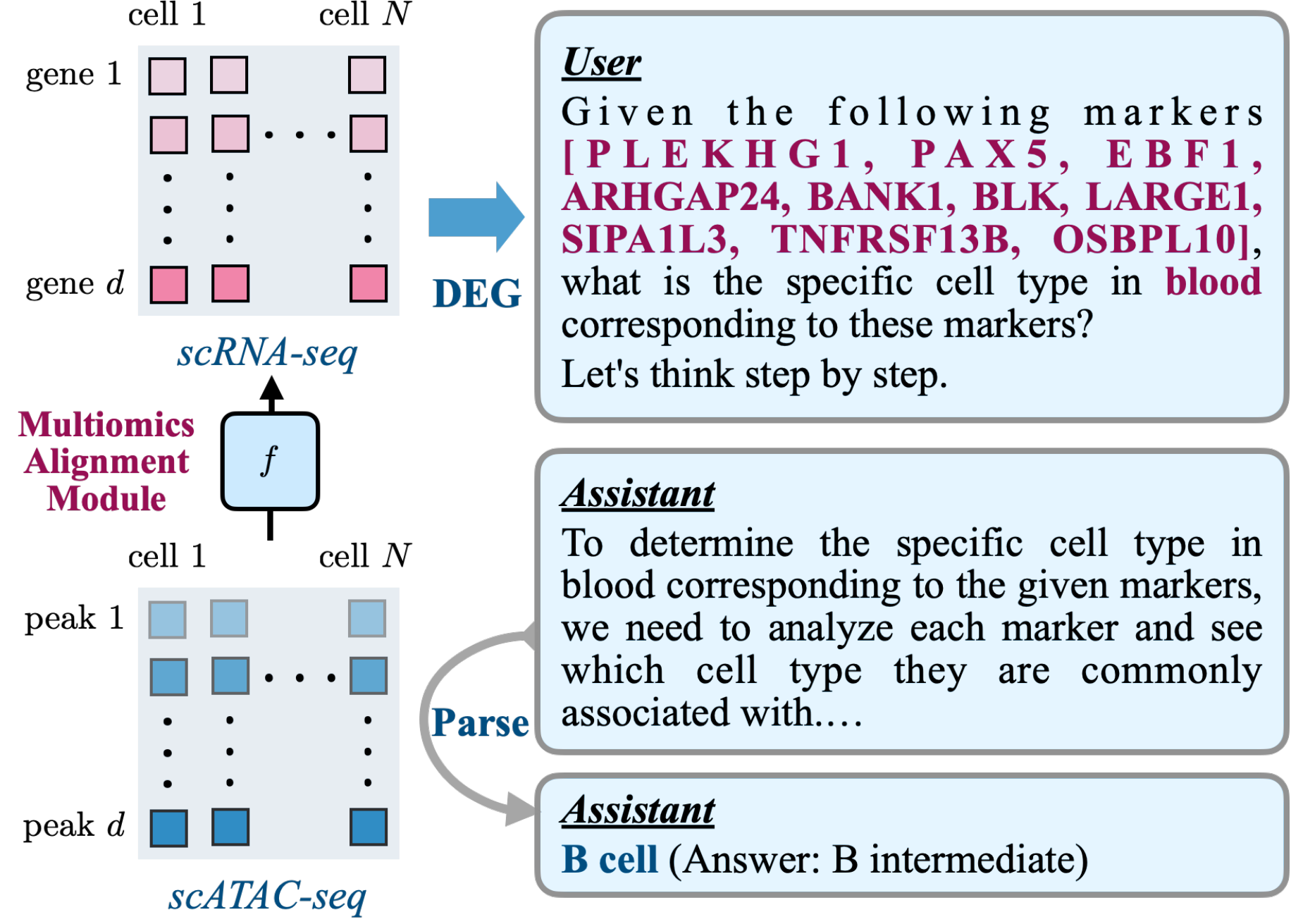}
         \caption{The illustration of prompting LLMs to finish the cell type annotation task.}
    \label{fig:prompt_illustration}
\end{figure}
\section{Method}
% As our objective is to assess the capability of current LLMs in analyzing single-cell biology data, we utilize instruction-tuned LLMs in this study without additional supervised fine-tuning on labeled data. 
We begin by introducing the foundational concepts of cell type annotation using scRNA-seq, followed by proposing two distinct zero-shot prompting strategies to guide the pre-trained LLMs in performing the cell type annotation task for single-cell analysis. Figure~\ref{fig:prompt_illustration} provides an overview of the proposed method.

\subsection{Preliminary of Cell Type Annotation}
Single-cell RNA sequencing (scRNA-seq) provides a detailed view of a cell's functionality by measuring the number of transcribed mRNA molecules (i.e., the number of expressed genes). Let a list of unique sequenced gene names be denoted as $\mathcal{G} = [g_1, \ldots, g_{d_g}]$. Consequently, a sequenced scRNA-seq sample can be represented as a continuous vector $\boldsymbol{x} \in \mathbb{R}^{d_g}$, where $\boldsymbol{x}_i$ denotes the observed expression value of the $i$th gene in the gene list $\mathcal{G}$. Based on the scRNA-seq data of a cell, denoted by $\boldsymbol{x}$, biologists can annotate its corresponding cell type, $\boldsymbol{y}$, by carefully analyzing gene co-expression patterns using domain expertise or by referencing established findings in the literature. 

Although cell type annotation is a crucial task for understanding the behavior of a cell, it demands significant labor and specialized domain knowledge. This raises a compelling question: ``Can contemporary large language models (LLMs) assist in analyzing single-cell genomics data?'' However, directly applying LLMs to this task poses significant challenges for several reasons: (1) Most state-of-the-art LLMs are trained exclusively on natural language, raising doubts about whether these models possess the same level of understanding and reasoning ability when applied to domain-specific content such as gene names and biological terminology, and (2) sequences obtained from many single-cell sequencing technologies cannot always be represented in conventional text formats suitable for LLM inputs, which limits the models' capacity to reason about various biological signals.

% In the following sections, we detail how these challenges can be addressed through the use of appropriate prompting strategies and cross-modality transformation modules.

\subsection{Zero-Shot Prompting Strategies for Cell Type Annotation}
\label{sec:promting_method}
To leverage LLMs for the task of cell type annotation, we formulate it as a standard question-answering problem. Specifically, we define the question $\boldsymbol{q}$ using the following template:
\textit{``Given the following markers $\{\texttt{X}\}$, what is the specific cell type in $\{\texttt{C}\}$ corresponding to these markers?''}
Here, $\{\cdot\}$ represents a placeholder, with $\texttt{X}$ denoting a gene expression profile description derived from an scRNA-seq vector $\boldsymbol{x}$, and $\texttt{C}$ representing any associated metadata for the cell (e.g., tissue type). To guide the LLMs towards generating an appropriate response for annotating cell types, we append a trigger sentence $\boldsymbol{t}$ after the question $\boldsymbol{q}$. Consequently, the LLMs are tasked with generating a response sentence $\boldsymbol{r}$ as follows:
\begin{align}
    \boldsymbol{r} = \textrm{LLM}(\boldsymbol{q}, \boldsymbol{t}).
    \label{eq:llm_prompt}
\end{align}

\paragraph{Gene Expression Profile Description} As previously mentioned, the scRNA-seq vector $\boldsymbol{x}$ is structured data, where each dimension represents the expression value of a specific gene. To enable LLMs to process this structured data, a serialization method must be employed. In our experiments, we used Differential Gene Expression (DGE) analysis to select the $k$ most distinguishable genes, $\{\hat{g}_1, \ldots, \hat{g}_k\}$, and then listed the selected $k$ gene names in decreasing order of their $p$-values, which can be expressed as follows:
\begin{align}
    \texttt{X} = [\hat{g}_1, \ldots, \hat{g}_k], ~~\hat{g}_i &\in \textrm{DEG}(\mathcal{G}, \boldsymbol{x}),
    \label{eq:gene_text}
\end{align}
where $\hat{g}_i$ represents the gene name of the $i$th selected differentially expressed gene, and $\texttt{X}$ is the formulated gene expression profile in text format to be provided as input to the LLMs.

\paragraph{Zero-Shot Prompting} To guide LLMs in annotating cell types based on the provided gene expression profile, we use a zero-shot prompting trigger sentence, $\boldsymbol{t}_{zero}$, which is set as: \textit{``The most likely cell type (directly return one cell type name) is''}. Given a cell $\boldsymbol{x}$ to be annotated and its gene expression profile $\texttt{X}$, the response $\boldsymbol{r}$ generated by the LLMs using (\ref{eq:llm_prompt}) is parsed to extract the predicted cell type, denoted as $\hat{\boldsymbol{y}}$.

\paragraph{Zero-Shot Chain-of-Thought Prompting} Cell type annotation requires reasoning skills to analyze complex co-expression patterns of different genes, which presents a significant challenge, even for expert biologists. Inspired by the chain-of-thought (CoT) method proposed by \citet{kojima2022large}, we adopted a two-stage strategy to enhance the reasoning capabilities of LLMs for cell type annotation. First, a chain-of-thought trigger sentence, $\boldsymbol{t}_{cot}$, is set as: \textit{"Let's think step by step."}. This prompts the LLMs to sequentially reason through the genes listed in $\texttt{X}$ one by one
\begin{align}
    \boldsymbol{z} = \textrm{LLM}(\boldsymbol{q}, \boldsymbol{t}_{cot}),
\end{align}
where $\boldsymbol{z}$ is the generated response from the first stage. In the second stage, the response $\boldsymbol{z}$, along with the initial prompt, is used to prompt the LLMs to summarize the final annotation using the trigger sentence $\boldsymbol{t}_{sum}$ as follows:
\begin{align}
    \boldsymbol{r} = \textrm{LLM}(\boldsymbol{q}, \boldsymbol{t}_{cot}, \boldsymbol{z}, \boldsymbol{t}_{sum}),
\end{align}
where $\boldsymbol{t}_{sum}$ is set as: \textit{"In summary, the most likely cell type (please return one cell type name) is"}. The final response is then parsed as the predicted cell type annotation $\hat{\boldsymbol{y}}$.

\subsection{Multiomics Cell Type Annotation}
Most multiomics sequencing signals cannot be directly expressed as text descriptions, which limits the ability of LLMs to interpret such data across different modalities in the same way they can process RNA-seq data. However, multiomics sequencing technologies provide complementary information, enabling the profiling of molecular quantities within cells across different biological layers. For example, ATAC-seq (Assay for Transposase-Accessible Chromatin using sequencing) is a widely used technique that measures DNA accessibility within the nucleus~\cite{wu2021babel}. In higher eukaryotes, DNA is typically in a highly compact and inaccessible state, but certain regions of the chromosome can become accessible in a cell-type-specific manner to perform functions such as transcription regulation. ATAC-seq assesses chromatin accessibility in individual cells, producing a sparse, binary matrix, where a value of 1 indicates an accessible region, and 0 indicates an inaccessible region. For brevity, we use ATAC-seq as a demonstration of how LLMs can be applied to solve the multiomics cell type annotation task.

\paragraph{Multiomics Alignment Module} To address this challenge, we propose a multiomics alignment method that enables LLMs to reason across different multiomics biological sequences. Specifically, given the strong natural language processing capability of LLMs, we select RNA-seq as the pivot modality for aligning various types of multiomics data. The data from different modalities are aligned with RNA-seq using a multi-modal alignment module, $f: \mathbb{R}^{d^\prime} \rightarrow \mathbb{R}^{d}$, which maps the ATAC-seq modality $\boldsymbol{x}^\prime \in \mathbb{R}^{d^\prime}$ into the RNA-seq modality $\boldsymbol{x}$, where $\boldsymbol{x} = f(\boldsymbol{x}^\prime)$. In our experiments, a variational auto-encoder~\cite{kingma2013auto} is used as the multi-modal alignment module $f$ to align both RNA-seq and ATAC-seq data. The details of the pre-trained VAE are provided in Appendix~\ref{app:cross-modality}. Once training is complete, the cross-modality alignment result $f(\boldsymbol{x}^\prime)$ is then transformed into a textual description using (\ref{eq:gene_text}). Finally, the LLM is provided with the prompted question $\boldsymbol{q}$ and the corresponding trigger $\boldsymbol{t}$ as input, generating sentences that are parsed to produce the final annotation $\hat{\boldsymbol{y}}$.
\begin{table*}[t]
    \centering
    \resizebox{0.78\textwidth}{!}{
    \begin{tabular}{lcccccc}
        \toprule
         & \multicolumn{3}{c}{\textbf{Zero-Shot}} & \multicolumn{3}{c}{\textbf{Zero-Shot CoT}}\\
        \textbf{Model}    & \textbf{BLEU-1}   & \textbf{BLEU-2} & \textbf{Average} & \textbf{BLEU-1} & \textbf{BLEU-2} & \textbf{Average} \\
        % \midrule
        \midrule
DeepSeek-LLM-67B  & 28.27 & 10.07 & 16.87 & 33.72 & 13.10 & 21.02 \\
Qwen2-72B     & 18.59 & 6.67  & 11.13 & 36.85 & 17.92 & 25.69 \\
Llama-3-70B    & 23.21 & 8.85  & 14.33 & 25.94 & 11.64 & 17.38 \\
Mixtral-8$\times$7B  & 16.94 & 6.17  & 10.23 & 31.57 & 13.83 & 20.90 \\
Mixtral-8$\times$22B & 26.42 & 10.49 & 16.65 & 40.96 & 19.40 & 28.19 \\
Cell2Sentence          & 25.24 & 11.79 & 17.25 & 25.24 & 11.79 & 17.25 \\\midrule
GPT-4o mini            & 45.74 & 23.29 & 32.64 & 50.29 & 27.89 & 37.45 \\
GPT-4o                 & 62.85 & 42.68 & 51.79 & 55.27 & 32.15 & 42.15 \\
        
        \bottomrule
    \end{tabular}
    }
    \caption{The BLEU evaluation of cell type annotation results on the \benchmarkName{} benchmark using the zero-shot and zero-shot chain-of-thought (CoT) prompting strategies to prompt LLMs respectively.}
    \label{tab:zero_shot_benchmark}
\end{table*}

\section{Benchmark Setup}
% In this section, we provide the detailed descriptions of the \benchmarkName{} and \benchmarkMulName{} datasets, which are designed to evaluate the reasoning capabilities of LLMs in various modalities of single cell data.

\subsection{\benchmarkName{}: Cell Type Annotation Benchmark on scRNA-seq data}
\benchmarkName{} is a benchmark dataset designed to evaluate the reasoning capabilities of LLMs in the field of single-cell genomics. Following previous work~\cite{hou2024assessing}, we curated cell type annotation samples from the following single-cell RNA sequencing datasets: Azimuth~\cite{hubmap2019human}, Human Cell Landscape (HCL)~\cite{han2020construction}, Mouse Cell Atlas (MCA)~\cite{han2018mapping}, GTEx~\cite{eraslan2022single}, B-cell lymphoma (BCL)~\cite{liu2023single}, Literature~\cite{eraslan2022single}, Colon Cancer~\cite{lee2020lineage}, Lung Cancer~\cite{kim2020single}, Tabula Sapiens (TS)~\cite{the2022tabula}, and Non-model Mammal~\cite{chen2021single}. A brief summary of all datasets utilized in this work is provided in Table~\ref{tab:dataset_stat}.

\paragraph{DEG Analysis and Cell Type Normalization} For each scRNA-seq, manually annotated cell types ($\boldsymbol{y}$) and gene expression matrices ($\boldsymbol{x}$) were obtained directly from the corresponding publications. For differential gene expression (DEG) analysis, raw gene expression counts were first log-transformed to the total sum of the maximum gene counts after adding a pseudocount of 1, using SCANPY~\cite{wolf2018scanpy}. Welch's t-test was then performed to identify differentially expressed genes (DEGs) by comparing each cell type against all others. Genes for each cell type were ranked in ascending order based on $p$-values, with ties in $p$-values further ranked in descending order by t-statistics. The top 10 DEGs were selected to construct the gene list $\texttt{X}$. To account for synonymy in cell type descriptions, we normalized cell type annotations to their unambiguous cell ontology (CL) names~\cite{jupp2015new}, using the API\footnote{https://bioportal.bioontology.org/} provided by~\citet{noy2009bioportal}. All synonyms for a cell type name returned by the API were included as ground truth annotation candidates ($\boldsymbol{y}$).

After data pre-processing, \benchmarkName{} contains \textbf{1191} cell types spanning \textbf{37} different tissues. Detailed statistics on tissue distribution, along with the complete tissue list, are illustrated in Figure~\ref{fig:nature_tissue_distribution}.

\subsection{\benchmarkMulName{}: Cell Type Annotation Benchmark on Multiomics Data}
\benchmarkMulName{} is a benchmark dataset designed to evaluate the reasoning capabilities of LLMs on single-cell multiomics data. For this dataset, we included publicly available single-cell multiomics data, specifically the human peripheral blood mononuclear cells (PBMC) 10k dataset from 10X Genomics, as well as the prefrontal cortex (PFC) brain dataset~\cite{emani2024single}, both of which feature parallel scRNA-seq and scATAC-seq sequencing. The pre-processing details are provided in Appendix~\ref{app:data_preprocessing_for_multiomics}. The DEG analysis and cell ontology name normalization were conducted following the same procedures as in \benchmarkName{}. After the data pre-processing stage, \benchmarkMulName{} contains 28 cell types for the human peripheral blood mononuclear cells (PBMC) 10k dataset, and 7 major cell types (excitatory neurons, inhibitory neurons, astrocytes, endothelial cells, microglia, oligodendrocytes, and OPCs) for the prefrontal cortex (PFC) brain dataset.

\begin{figure*}[t]
    \centering
    \begin{subfigure}[b]{0.8\textwidth}
         \centering
         \includegraphics[width=\textwidth]{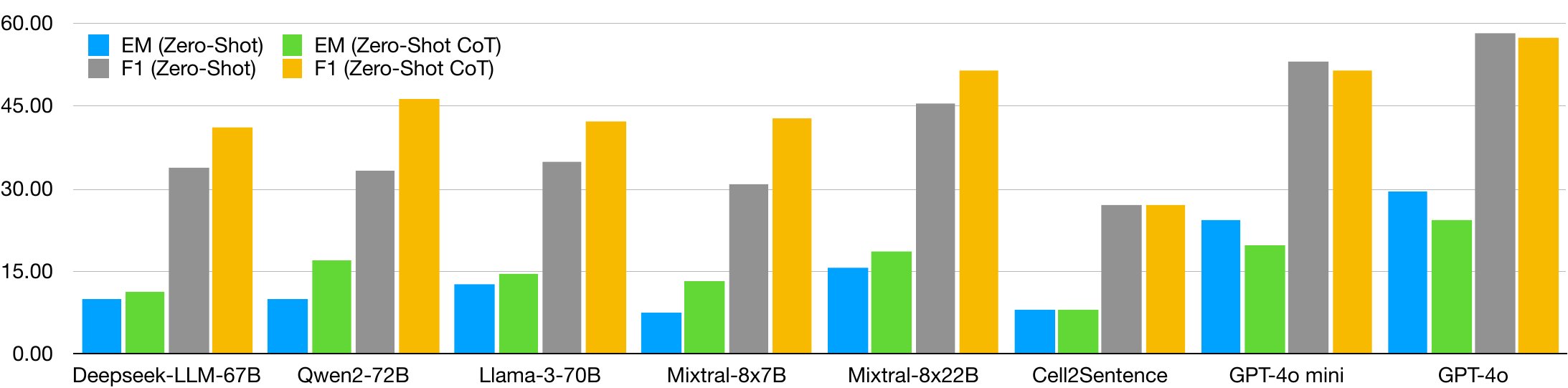}
         % \label{fig:y equals x}
     \end{subfigure}
\caption{The EM and F1 evaluation results on the \benchmarkName{} benchmark using two zero-shot prompting strategies.}
    \label{fig:nature_method_squad}
\end{figure*}

\section{Experiment Results}

\subsection{Evaluation Metrics}
Cell type annotation is provided in free format, without a strict answer template, which differs from multiple-choice or binary classification tasks. Additionally, cell type annotation often involves multiple synonyms, requiring the evaluation of the predicted answer against several candidates. To fairly assess the free-format responses for cell type annotation, we propose using the BLEU score~\cite{Papineni02bleu:a,lin-och-2004-orange} to measure the n-gram overlap between the predicted text $\hat{\boldsymbol{y}}$ and the ground truth annotation $\boldsymbol{y}$. Specifically, we report BLEU-1, BLEU-2, and the geometric average BLEU score, as many cell type annotations are fewer than 3-grams in length. To further evaluate the precision of cell type annotation generated by LLMs, we also employ EM and F1, metrics commonly used to assess the quality of responses in question-answering tasks~\cite{Rajpurkar2016SQuAD10}.

\begin{figure}[t]
    \centering
    \begin{subfigure}[b]{0.4\textwidth}
         \centering
         \includegraphics[width=\textwidth]{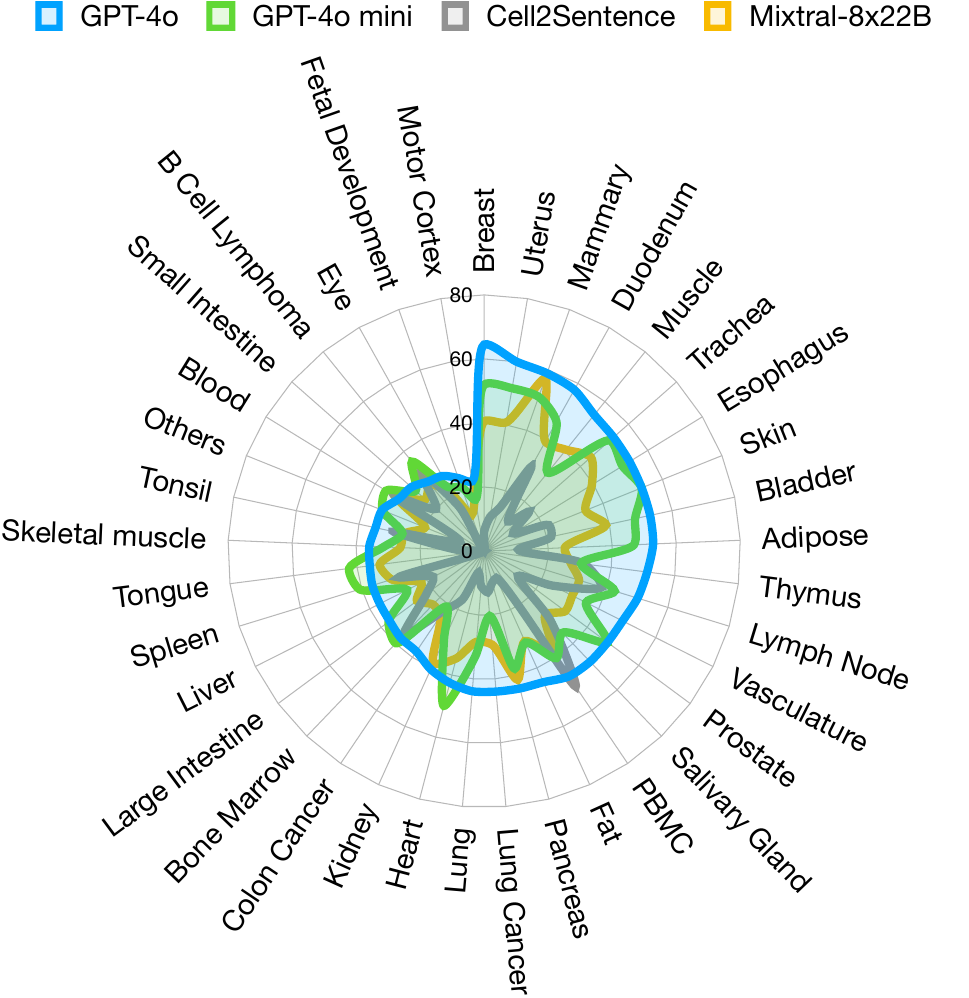}
         % \caption{Zero-Shot Prompting Strategy}
         % \label{fig:y equals x}
     \end{subfigure}
     % \hfill
     \\
    \begin{subfigure}[b]{0.4\textwidth}
         \centering
         \includegraphics[width=\textwidth]{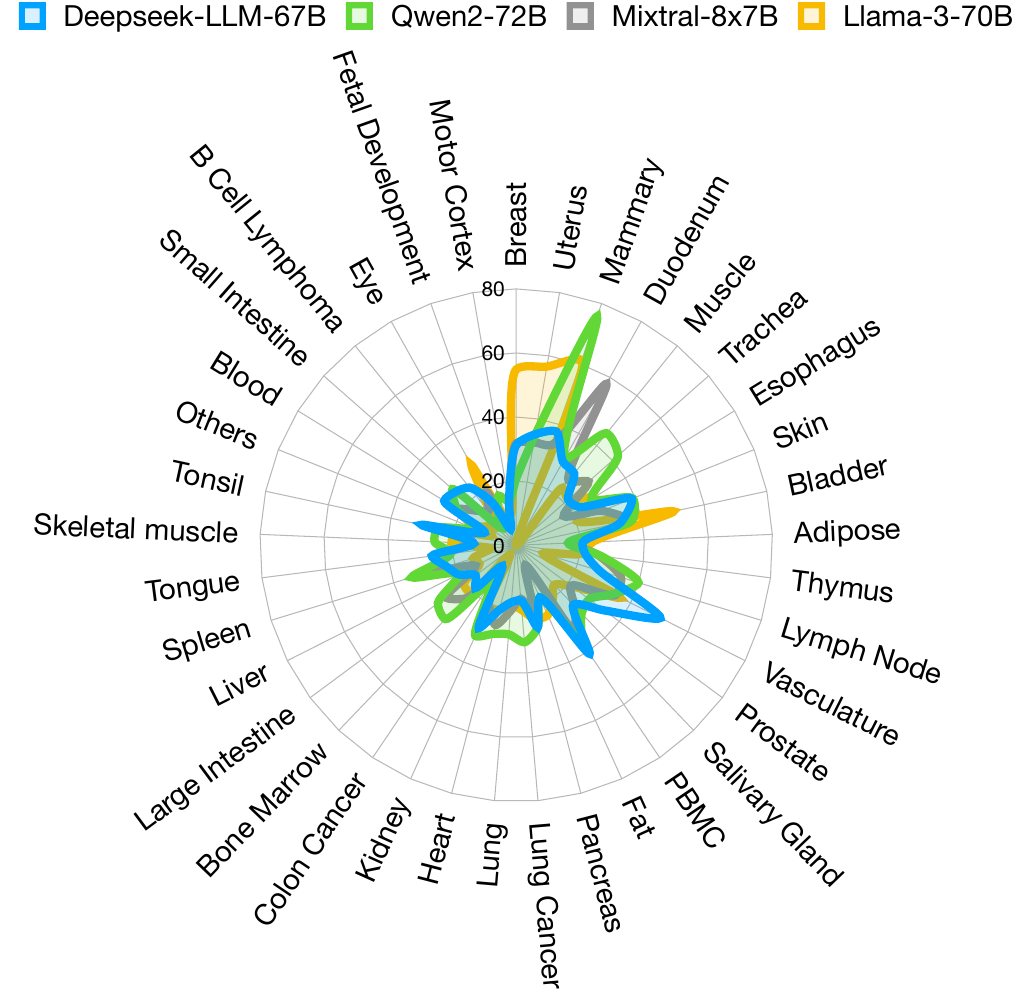}
         % \caption{Zero-Shot Chain-of-Thought Prompting Strategy}
         % \label{fig:y equals x}
     \end{subfigure}
\caption{The BLEU evaluation result per tissue of the \benchmarkName{} benchmark.}
    \label{fig:nature_method_bleu_per_tissue}
\end{figure}

\subsection{Large Language Models}
In this work, we conducted detailed evaluations of single-cell cell type annotation ability across both open-source and close-source instruction-tuned LLMs, such as Qwen2~\cite{yang2024qwen2} Llama-3~\cite{dubey2024llama}, Mixtral~\cite{jiang2024mixtral}, DeepSeek~\cite{deepseek-aiDeepSeekLLMScaling2024}, and GPT-4o~\cite{achiam2023gpt}. To further compare the reasoning ability of general instructed and domain-specific LLMs, we also include Cell2Sentence~\cite{levinecell2sentence} in our benchmark. A brief summary of all LLMs utilized in this work is provided in Table~\ref{tab:llms}. The implementation details are described in Appendix~\ref{app:implementation_details}.

% \input{latex/table/llms}

% \paragraph{Implementation Details} For our experiments, we used a consistent set of hyperparameters across all instruction-tuned models, adhering to commonly accepted values from previous studies~\cite{wolf2019huggingface}. Specifically, we set the temperature at 0.6, top\_p at 0.9, and top\_k at 50. 
% The temperature parameter adjusts the randomness during sampling, top\_p filters out tokens with lower probabilities, and top\_k limits the sampling process to the $k$ most probable tokens. 
% The detailed method for cleansing and extracting the annotation result from the response is described in Appendix~\ref{app:answer_cleansing}. For Cell2Sentence, we followed the instruction finetuning method proposed in~\cite{levinecell2sentence} to train the pretrained Cell2Sentence model using the instruction-following dataset provided by the authors. Our code is available in the Supplementary Materials.

\begin{table*}[t]
    \centering
    \resizebox{0.78\textwidth}{!}{
    \begin{tabular}{lcccccc}
        \toprule
         & \multicolumn{3}{c}{\textbf{RNA-seq}} & \multicolumn{3}{c}{\textbf{ATAC-seq}}\\
        \textbf{Model}    & \textbf{BLEU-1}   & \textbf{BLEU-2} & \textbf{Average} & \textbf{BLEU-1} & \textbf{BLEU-2} & \textbf{Average} \\
        % \midrule
        \midrule
% DeepSeek-LLM-67B  & 6.20  & 0.42  & 1.61  & 8.79  & 1.69  & 3.85  \\
Qwen2-72B      & 20.77 & 5.32  & 10.51 & 16.07 & 3.79  & 7.80  \\
Llama-3-70B     & 27.41 & 10.10 & 16.64 & 29.84 & 10.23 & 17.47 \\
Mixtral-8$\times$7B   & 33.09 & 18.45 & 24.71 & 23.08 & 9.77  & 15.02 \\
Mixtral-8$\times$22B  & 20.63 & 8.06  & 12.90 & 20.24 & 8.33  & 12.99 \\
Cell2Sentence          & 41.35 & 35.29 & 38.20 & 36.04 & 22.67 & 28.58 \\\midrule
GPT-4o mini            & 37.40 & 21.05 & 28.06 & 35.46 & 17.14 & 24.66 \\
GPT-4o                 & 43.75 & 26.32 & 33.93 & 38.84 & 21.18 & 28.68 \\
        
        \bottomrule
    \end{tabular}
    }
    \caption{The BLEU evaluation results of cell type annotation on the \benchmarkMulName{} benchmark using the zero-shot chain-of-thought (CoT) prompting strategies to prompt LLMs.}
    \label{tab:scact_benchmark}
\end{table*}

% \subsection{Evaluations of Cell Type Annotations on \benchmarkName{}} 
\subsection{Evaluations on \benchmarkName{}} 
We first evaluated the cell type annotation capabilities of LLMs on our proposed \benchmarkName{}. The BLEU scores for \benchmarkName{} are reported in Table~\ref{tab:zero_shot_benchmark}. As shown in the table, we observed that general open-source LLMs, such as Mixtral-8$\times$22B using the zero-shot prompting strategy, achieved performance comparable to the domain-specific pretrained model (i.e., Cell2Sentence). Specifically, the average BLEU score for Mixtral-8$\times$22B was 16.65, compared to 17.25 for Cell2Sentence. After applying zero-shot chain-of-thought (CoT) prompting, all open-source LLMs and GPT-4o mini showed significant improvement in BLEU scores. For instance, the average BLEU score of Mixtral-8$\times$22B using zero-shot CoT increased to 28.19, representing a relative improvement of over 69\% compared to the zero-shot prompting strategy. This result demonstrates the effectiveness of the zero-shot CoT prompting method in enhancing LLM performance for cell type annotations.

Furthermore, we observed that the zero-shot CoT strategy enabled all open-source LLMs, which were trained solely on natural language, to significantly outperform the domain-specific pretrained model in cell type annotation tasks. For instance, the best result achieved by an open-source LLM using zero-shot CoT (i.e., Mixtral-8$\times$22B) surpassed the domain-specific pretrained model (i.e., Cell2Sentence) by over 64\% in relative performance improvement. As shown in Figure~\ref{fig:nature_method_squad}, a similar trend was evident in the evaluation results for EM and F1 scores. Open-source LLMs demonstrated superior performance in both EM and F1 metrics compared to the domain-specific model when applying the zero-shot CoT prompting strategy described in Section~\ref{sec:promting_method}. For example, Qwen2-72B achieves approximately a 35\% relative improvement in F1 score when using the zero-shot CoT prompting strategy (46.38) compared to the standard zero-shot prompting approach (33.15). Given the free-form nature of cell type annotation tasks and the abundance of synonyms and word permutations used to describe a single cell type, even after normalizing the annotations using unambiguous cell ontology (CL) names, the EM metric remains overly restrictive, leading to lower performance scores.

These observations highlight the reasoning capabilities of LLMs in analyzing complex gene co-expression patterns and their broad acquisition of domain-specific knowledge. This underscores the potential of LLMs trained on text domains to outperform domain-specific models pretrained from scratch in specialized tasks. Consequently, it motivates the research community to focus more on leveraging the domain-specific knowledge and reasoning abilities inherent in existing LLMs, which may offer superior problem-solving capabilities in specialized domains.

Nevertheless, it is important to note that there remains a significant performance gap between open-source LLMs and closed-source models (i.e., GPT-4o mini and GPT-4o) across all metrics as shown in Table~\ref{tab:zero_shot_benchmark} and Figure~\ref{fig:nature_method_squad}, including BLEU score and EM/F1 performance. Additionally,  we found that applying zero-shot CoT prompting slightly degraded the cell type annotation performance of GPT-4o in F1 score (58.29 vs 57.36) as shown in Figure~\ref{fig:nature_method_squad}. A potential explanation for this is that GPT-4o is already fine-tuned for chain-of-thought reasoning, even when users do not explicitly prompt it to do so~\cite{chen2023you}. Based on our observations, using zero-shot CoT may cause GPT-4o to generate broader cell type names rather than more specific sub-cell type names, which negatively impacts precision and recall scores.

% \textcolor{red}{I don't quite understand this following para}
% \paragraph{Evaluation Results on Tissues} 
To better understand the generalization ability of LLMs in annotating cell types across different tissues, we analyzed the annotation accuracy for each tissue. The average BLEU score per tissue is presented in Figure~\ref{fig:nature_method_bleu_per_tissue}. For the open-source Mixtral-8$\times$22B, its performance closely aligns with the GPT-4o series across various tissues, such as breast, mammary, and pancreas, likely due to the highly correlated language corpus used during the pretraining stage.
For well-studied tissues with extensive literature, such as peripheral blood mononuclear cells (PBMC), duodenum, breast, uterus, and mammary glands, large language models (LLMs) tend to exhibit stronger performance in reasoning and accurately identifying corresponding cell types as shown in Figure~\ref{fig:nature_method_bleu_per_tissue}. Moreover, we observed that the domain-specific model (i.e., Cell2Sentence) does not demonstrate as consistent performance across various tissues compared to general LLMs. This finding further supports the ability of LLMs to reason over complex gene co-expression patterns and reflects the broad domain-specific knowledge they acquire during training. Detailed EM/F1 evaluation results (Appendix~\ref{app:results_on_tissues}) corroborate these conclusions.
% For well-studied tissues with extensive literature, such as PBMC, duodenum, breast, uterus, and mammary glands, LLMs tend to perform better at reasoning and identifying corresponding cell types \textcolor{red}{cite fig/table}. Furthermore, we observed that the domain-specific model (i.e., Cell2Sentence) does not achieve as balanced performance across different tissues when compared to general LLMs. This finding further proved the LLMs' ability to reason over complex gene co-expression patterns and the broad domain-specific knowledge they acquire during training. Detailed EM/F1 evaluation results, provided in Appendix~\ref{app:results_on_tissues}, reveal similar conclusions.

\begin{figure}[t]
    \centering
    \begin{subfigure}[b]{0.4\textwidth}
         \centering
         \includegraphics[width=\textwidth]{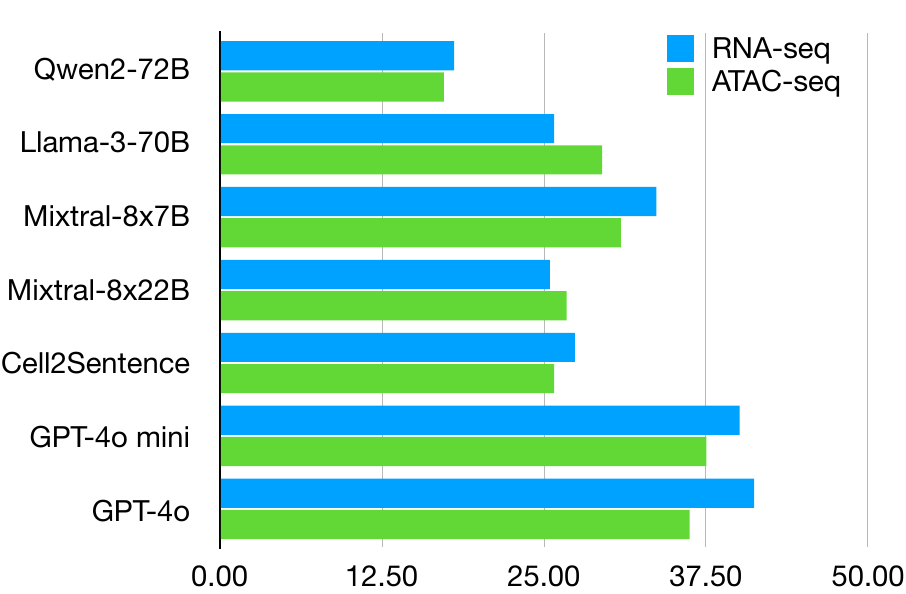}
         % \caption{Zero-Shot Prompting Strategy}
         % \label{fig:y equals x}
     \end{subfigure}
\caption{The F1 evaluation results on the \benchmarkMulName{} benchmark.}
    \label{fig:scact_f1}
\end{figure}

% \subsection{Evaluations of Cell Type Annotation on \benchmarkMulName{}}  
\subsection{Evaluations on \benchmarkMulName{}}  
To assess whether LLMs can effectively analyze multiomics data, we further evaluated cell type annotation performance using the proposed \benchmarkMulName{} benchmark. Following the methodology described earlier, for RNA-seq data, we directly formatted the gene expression vector $\boldsymbol{x}$ into a profile description $\texttt{X}$. For ATAC-seq data, we first pretrained a cross-modality alignment model, $f$, using a variational autoencoder (VAE) architecture to align both RNA-seq and ATAC-seq data into a common semantic space. Consequently, the ATAC-seq vector $\boldsymbol{x}^\prime$ was translated into the RNA-seq domain using $f(\boldsymbol{x}^\prime)$, and subsequently formatted into a profile description $\texttt{X}$. With this unified semantic space, we can bridge any omic data with RNA-seq, enabling multiomics data to be processed by LLMs through pure text inputs. The translation performance of the pretrained VAE is reported in Appendix~\ref{app:cross-modality}, demonstrating a strong alignment between RNA-seq and ATAC-seq data.

We summarize the multiomics cell type annotation results in Table~\ref{tab:scact_benchmark} and Figure~\ref{fig:scact_f1}. As shown in the results, GPT-4o, GPT-4o mini, and Mixtral-8$\times$22B achieve comparable annotation performance to the domain-specific model Cell2Sentence on both RNA-seq and ATAC-seq data. For example, GPT-4o achieves a BLEU score of 33.93 and an F1 score of 41.28, compared to Cell2Sentence’s BLEU score of 38.20 and F1 score of 27.41. This observation aligns with the tissue-level annotation results shown in Figure~\ref{fig:nature_method_bleu_per_tissue}, where the majority of samples in \benchmarkMulName{} come from PBMC tissue, which is well annotated by the Cell2Sentence model. As illustrated in Figure~\ref{fig:scact_f1}, the comparable performance of cell type annotation from ATAC-seq data demonstrates the ability of LLMs to analyze multiomics data using the multimodal alignment module $f$. This finding highlights the potential for adapting LLMs to analyze a broader range of biological sequencing data.

% \paragraph{Impact of Alignment Module $f$}
% \junhao{discuss the result of gene score and VAE}

\section{Conclusions}
This study introduces \paperacronym{}, a pioneering benchmarking effort that evaluates the capabilities of instruction-tuned large language models (LLMs) for cell type annotation across single-cell genomics data from various modalities. By curating a diverse dataset encompassing multiple species and cell types, we systematically assessed the ability of LLMs to process and analyze complex biological data. Our findings demonstrate that LLMs exhibit strong interpretive capabilities in scRNA-seq data, even without extensive fine-tuning, and can effectively generate reasoning processes that support biological insights through the use of chain-of-thought (CoT) prompting techniques. Additionally, we explored the application of these models to multi-omics data, highlighting their potential for cross-modality analysis and providing a foundation for future advancements in automated single-cell annotation. This work underscores the promise of LLMs in transforming cell type annotation workflows, while also emphasizing the need for ongoing innovation to fully exploit their potential across diverse molecular modalities.
\section*{Limitations}
Although extensive evaluations have been conducted on the proposed benchmarks, \benchmarkName{} and \benchmarkMulName{}, further evaluation is needed to fully understand the ability of LLMs to analyze genomics data. One area for improvement involves a thorough investigation into retrieval-augmented generation (RAG) during the reasoning stage of cell type annotation. This approach aims to enhance LLMs' reasoning capabilities by leveraging existing biological knowledge, thereby improving the precision of annotating novel and complex gene co-expression patterns. Additionally, due to data constraints, the majority of multiomics benchmarks in this study are limited to RNA-seq and ATAC-seq data. While the proposed benchmark has significant influence within the large language model for scientific data community, comprehensive validation across a broader range of omics data is essential to fully assess LLMs' capabilities in analyzing biological data. These improvements can be facilitated through collaborative efforts within the research community to collect more publicly accessible multiomics data. By pursuing these research directions, we anticipate further refinement and expansion of the scope of our benchmarks.

\bibliography{main}

\clearpage
\appendix
% \section{Prompting Example}
% \label{app:prompting-example}

\section{Answer Cleansing}
\label{app:answer_cleansing}
Since the cell type annotation is in free format, we extract the final answer by matching any character that is not a newline (\textbackslash n), comma (,), or period (.), zero or more times. The pseudo code to implement this is \verb|pred = re.match(r"^[^\n,\.]*", pred)|. The first matched result is selected as the cell type annotation. The annotation is further normalized to its singular form for comparison\footnote{https://github.com/jaraco/inflect}. 

\begin{figure*}[t]
    \centering
    \begin{subfigure}[b]{0.9\textwidth}
         \centering
         \includegraphics[width=\textwidth]{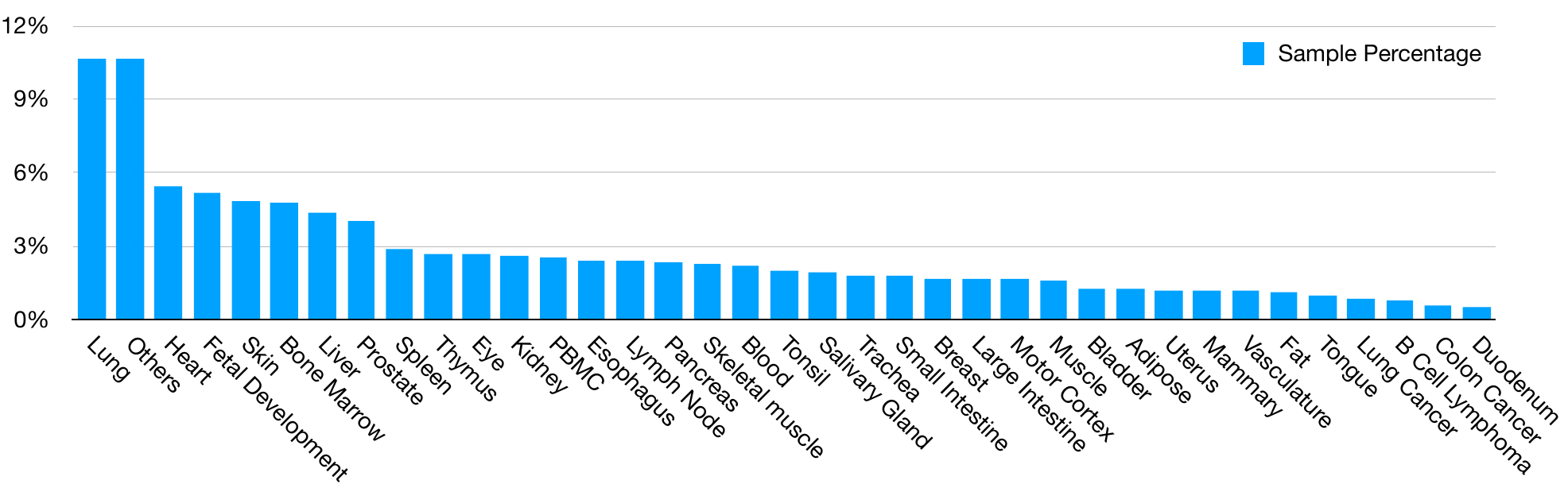}
         % \caption{Zero-Shot Prompting Strategy}
         % \label{fig:y equals x}
     \end{subfigure}
\caption{The statistics results of percentages of tissue in the \benchmarkName{} dataset.}
    \label{fig:nature_tissue_distribution}
\end{figure*}

\section{Dataset Statistics}
\label{app:dataset}
For \benchmarkName{}, We curated cell type annotation samples from the following single-cell RNA sequencing datasets: Azimuth~\cite{hubmap2019human}, Human Cell Landscape (HCL)~\cite{han2020construction}, Mouse Cell Atlas (MCA)~\cite{han2018mapping}, GTEx~\cite{eraslan2022single}, B-cell lymphoma (BCL)~\cite{liu2023single}, Literature~\cite{eraslan2022single}, Colon Cancer~\cite{lee2020lineage}, Lung Cancer~\cite{kim2020single}, Tabula Sapiens (TS)~\cite{the2022tabula}, and Non-model Mammal~\cite{chen2021single}. For \benchmarkMulName{}, we included publicly available single-cell multiomics data, specifically the human peripheral blood mononuclear cells (PBMC) 10k dataset from 10X Genomics, as well as the prefrontal cortex (PFC) brain dataset~\cite{emani2024single}, both of which feature parallel scRNA-seq and scATAC-seq sequencing. Table~\ref{tab:dataset_stat} provides detailed information about \benchmarkName{} and \benchmarkMulName{}.

\begin{table*}[!htbp]
    \centering
    \resizebox{0.9\textwidth}{!}{
    \begin{tabular}{llr}
        \toprule
        \textbf{Benchmark}       & \textbf{Dataset}  & \textbf{\# Sample}\\
        % \midrule
        \midrule
        \multirow{9}{*}{\benchmarkName{}} &  Azimuth~\cite{hubmap2019human}  & 324   \\
               & Human Cell Landscape~\cite{han2020construction}  & 62  \\
               &  Mouse Cell Atlas (MCA)~\cite{han2018mapping}   & 65   \\
          &  GTEx~\cite{eraslan2022single}  & 99\\
          & Literature~\cite{eraslan2022single}  & 90\\
          &  B-cell lymphoma (BCL)~\cite{liu2023single}   & 9 \\
           &  Colon Cancer~\cite{lee2020lineage}  & 7 \\
           & Lung Cancer~\cite{kim2020single}  & 10\\
           & Tabula Sapiens (TS)~\cite{the2022tabula} & 471 \\
           & Non-model Mammal~\cite{chen2021single}  & 54\\
        \midrule
        \multirow{2}{*}{\benchmarkMulName{}}   &  Human peripheral blood mononuclear cells (PBMC) 10k dataset & 28\\
           &  Prefrontal cortex (PFC) brain dataset~\cite{emani2024single} &  7\\
        \bottomrule
    \end{tabular}
    }
    \caption{The dataset statistics of proposed \paperacronym{}.}
    \label{tab:dataset_stat}
\end{table*}

\begin{table}[!htbp]
    \centering
    \resizebox{0.48\textwidth}{!}{
    \begin{tabular}{lrc}
        \toprule
        \textbf{Language Model}       & \textbf{Params} & \textbf{Instruction Format} \\
        % \midrule
        \midrule
        DeepSeek-LLM   &  67B & General Text  \\
        Qwen2       & 72B & General Text  \\
        Llama-3       &  70B & General Text    \\
        Mixtra-8$\times$7B  &  56B & General Text  \\
        Mixtra-8$\times$22B  &  176B & General Text  \\
        Cell2Sentence  &  160M & Cell Sentence  \\ \midrule
        GPT-4o mini   &  $\sim$8B & General Text  \\
        GPT-4o   &  $\sim$200B & General Text  \\
        % \midrule
        %  & Collombat, Patrick, et al.        & Collombat, Patrick, et al. & Ma, Qing, et al. \\
        \bottomrule
    \end{tabular}
    }
    \caption{The large language models utilized in our experiments.}
    \label{tab:llms}
\end{table}

\section{Data Pre-processing for \benchmarkMulName{}}
\label{app:data_preprocessing_for_multiomics}
\paragraph{scRNA-seq Data Pre-processing} For the scRNA-seq dataset, cells with fewer than 200 reads or suspected multiplets~\cite{doubletdetection} were filtered out. We then selected the top 3000 highly variable genes to construct the scRNA-seq matrix using Pegasus and Doubletdetection. Log-normalization was applied to the entire matrix $\boldsymbol{x}$, resulting in the final dataset for further annotation. To visualize the cell distribution, we performed LEIDEN clustering~\cite{leiden} based on PCA results reduced to 20 dimensions. Cell types were manually annotated using previously identified marker genes.

\paragraph{scATAC-seq Data Pre-processing} Similarly, cells with insufficient TSS enrichment (TSS score < 2.0), low sequencing depth (< 1000 reads), or suspected multiplets were filtered out using ArchR with default parameters~\cite{archr}. Peak calling was performed using MACS2~\cite{macs2}, and the TF-IDF algorithm~\cite{tfidf} was applied to retain the 100000 most informative peaks. The binarized matrix was then used for training the alignment module $f$ and for the cell type annotation process.

\section{Multiomics Alignment Module}
\label{app:cross-modality}

A single-cell multiomics dataset consists of $N$ single-cell multi-modal data points $\mathcal{C} = \{\boldsymbol{c}^{(1)}, \ldots, \boldsymbol{c}^{(N)}\}$, where each cell $\boldsymbol{c}^{(i)} = (\mathbf{x}^{(i)}, \mathbf{y}^{(i)})$ includes an ATAC-seq vector $\mathbf{x}^{(i)}$ and its corresponding RNA-seq vector $\mathbf{y}^{(i)}$, along with a semantic label $\ell^{(i)}$ indicating its cell type among $T$ classes. We can train the Multiomics Alignment Module $f$ by mapping RNA and ATAC modalities into the same embedding space $\mathcal{E}$. Please note that, to accurately describe the biological relationships, we use different notations for ATAC-seq and RNA-seq in this section compared to the main content.
% Briefly, this multiomics alignment model comprises two parts, as shown in \textbf{Figure~\ref{fig:method_overview}}: (1) a modality-specific autoencoder to pretrain each modality; (2) cross-modality alignment using adversarial training.
The advantage of using the embedding model is that it allows us to easily extend our multiomics alignment model, which learns from paired RNA and ATAC data with known paired cell type labels, to unpaired data without prior knowledge of cell type class labels.

% \paragraph{Modality-Specific AutoEncoder}
We adopt two autoencoders to model the modality-specific feature. For ATAC-seq, each dimension in $\mathbf{x}$ is considered a binary categorical feature, with one low-dimensional embedding for each category. The encoder projects the raw input into semantics features as
% $\mathbf{h}^{(i)}_a = f_{\textrm{Enc}}^a(\mathbf{W}^a_{\textrm{Emb}}(\mathbf{x}^{(i)})),~
%     \mathbf{h}^{(i)}_r = f_{\textrm{Enc}}^r(\mathbf{W}^r_{\textrm{Emb}}(\mathbf{y}^{(i)}))$,
\begin{equation}
    \begin{split}
        \mathbf{h}^{(i)}_a &= f_{\textrm{Enc}}^a(\mathbf{W}^a_{\textrm{Emb}}(\mathbf{x}^{(i)})),\\
        \mathbf{h}^{(i)}_r &= f_{\textrm{Enc}}^r(\mathbf{W}^r_{\textrm{Emb}}(\mathbf{y}^{(i)}))
    \end{split}
\end{equation}
where $\mathbf{W}^a_{\textrm{Emb}} \in \mathbb{R}^{d_h\times d_a}$ is a category embedding module to accommodate the high-dimensional ATAC-seq data, $\mathbf{W}^r_{\textrm{Emb}} \in \mathbb{R}^{d_h\times d_r}$ is an embedding matrix for RNA-seq, $f_{\textrm{Enc}}^a$ and $f_{\textrm{Enc}}^r$ are encoder networks to generate embeddings $\mathbf{h}_a, \mathbf{h}_r \in \mathbb{R}^{d_h}$ in $\mathcal{E}$ of dimension $d_h$. The decoder generates reconstructions via $\hat{\mathbf{x}}^{(i)} = f_{\textrm{Dec}}^a(\mathbf{h}^{(i)}_a),~\hat{\mathbf{y}}^{(i)} = f_{\textrm{Dec}}^r(\mathbf{h}^{(i)}_r)$,
% \begin{equation}
%     \begin{split}
%         \hat{\mathbf{x}}^{(i)} = f_{\textrm{Dec}}^a(\mathbf{h}^{(i)}_a),~
%         \hat{\mathbf{y}}^{(i)} = f_{\textrm{Dec}}^r(\mathbf{h}^{(i)}_r)
%     \end{split}
% \end{equation}
where $f_{\textrm{Dec}}^a$ and $f_{\textrm{Dec}}^r$ are two decoder networks for the two modalities, $\hat{\mathbf{x}}^{(i)}$ and $\hat{\mathbf{y}}^{(i)}$ represent the reconstructions with objective defined as
\begin{equation}
    % \mathcal{L}_{\textrm{Rec}} = \textstyle \sum_{i=1}^N \textrm{MSE}(\hat{\mathbf{x}}^{(i)}, \mathbf{x}^{(i)}) + \textrm{BCE}(\hat{\mathbf{y}}^{(i)}, \mathbf{y}^{(i)})
    \small \mathcal{L}_{\textrm{Rec}} = \mathbb{E}_{\boldsymbol{c}\sim\mathcal{C}}[\textrm{BCE}(\hat{\mathbf{x}}^{(i)}, \mathbf{x}^{(i)}) + \textrm{MSE}(\hat{\mathbf{y}}^{(i)}, \mathbf{y}^{(i)})]
\end{equation}
where $\textrm{BCE}$ is the binary cross-entropy loss, and $\textrm{MSE}$ is the mean-squared error.

\begin{figure*}[t]
    \centering
    \includegraphics[width=0.9\textwidth]{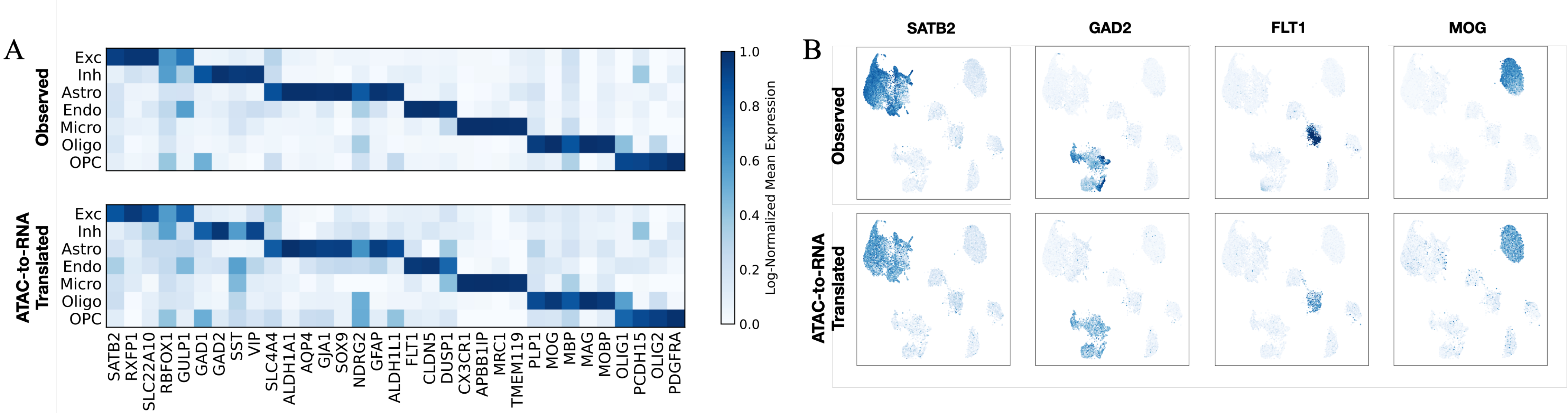}
    \caption{The trained multiomics alignment module $f$ can accurately predict the RNA-seq modality from given single-cell ATAC-seq profiles. (A) The comparison of predicted marker gene expression with actual values across different cell types demonstrated high consistency and specificity to cell type (mean $R^2$=0.914). (B) The UMAP of real scRNA-seq data, colored according to both actual and predicted expression levels for marker genes, exhibited a strong similarity.}
    \label{fig:fig3}
\end{figure*}

\paragraph{Alignment Embedding Adversarial Training} Given a cell type $T=k$, we define $\mathcal{C}^k$ as a subset of $\mathcal{C}$, where each cell $\boldsymbol{c}^{(i)} \in \mathcal{C}^k$ has the same label $\ell^{(i)} = k$. To align the modality-specific embeddings and capture the regulatory regulations between them, two mapping layers are adopted to jointly align the two modalities
% $\Tilde{\mathbf{h}}_{r}^{(i)} = f_{\textrm{AR}}(\mathbf{h}_a^{(i)}),~
%     \Tilde{\mathbf{h}}_{a}^{(i)} = f_{\textrm{RA}}(\mathbf{h}_r^{(i)})$,
\begin{equation}
    \begin{split}
        \Tilde{\mathbf{h}}_{r}^{(i)} = f_{\textrm{AR}}(\mathbf{h}_a^{(i)}),~
        \Tilde{\mathbf{h}}_{a}^{(i)} = f_{\textrm{RA}}(\mathbf{h}_r^{(i)})
    \end{split}
\end{equation}
where $f_{\textrm{AR}}$ aims to map the ATAC embeddings to the RNA embeddings and $f_{\textrm{RA}}$ does the opposite. We use a generative adversarial training mechanism~\citep{goodfellow2014generative} to let both encoders and mapping layers act as two generators to learn the modality-agnostic latent space $\mathcal{E}$. And then we apply the discriminator $D_{a}^k$ in each cell type $k$ for binary classification, aiming to differentiate whether $\mathbf{h}_a$ and $\Tilde{\mathbf{h}}_a$ of the ATAC embedding belongs to the cell type $k$ or not. The $D_{r}^k$ does the similar operation for the RNA embeddings $\mathbf{h}_r$ and $\Tilde{\mathbf{h}}_r$. Then, the discrimination loss can be formulated as
\begin{equation}
    \small
    \begin{split}
        \mathcal{L}_{\textrm{Dis}}^k = & \mathbb{E}_{\mathbf{x} \sim \mathcal{C}^k}[\log D_a^k(\mathbf{h}_a)] + \mathbb{E}_{\mathbf{y} \sim \mathcal{C}^k}[\log(1 - D_a^k(\Tilde{\mathbf{h}}_a))]  \\
        +                              & \mathbb{E}_{\mathbf{y} \sim \mathcal{C}^k}[\log D_r^k(\mathbf{h}_r)] + \mathbb{E}_{\mathbf{x} \sim \mathcal{C}^k}[\log(1 - D_r^k(\Tilde{\mathbf{h}}_r))].
    \end{split}
\end{equation}
The generators are trained to simultaneously fool the discriminator and keep the cycle consistency~\citep{zhu2020unpaired}
\begin{equation}
    \small
    \begin{split}
        \mathcal{L}_{\textrm{Gen}}^k = & \mathbb{E}_{\mathbf{x} \sim \mathcal{C}^k}[-\log D_r^k(\Tilde{\mathbf{h}}_r) + \textrm{MSE}(f_{\textrm{RA}}(\Tilde{\mathbf{h}}_r), \mathbf{h}_a)]  \\
        +                              & \mathbb{E}_{\mathbf{y} \sim \mathcal{C}^k}[-\log D_a^k(\Tilde{\mathbf{h}}_a) + \textrm{MSE}(f_{\textrm{AR}}(\Tilde{\mathbf{h}}_a), \mathbf{h}_r)]. \\
    \end{split}
\end{equation}
Therefore, the adversarial training process can be summarized in the following objective~function
\begin{equation}
    \mathcal{L}_{\textrm{Adv}} = \min_{\theta_{\textrm{Gen}}} \max_{\theta_{\textrm{Dis}}} \mathbb{E}_{k \sim T}[\mathcal{L}_{\textrm{Gen}}^k + \mathcal{L}_{\textrm{Dis}}^k]
\end{equation}
where $\theta_{\textrm{Gen}}$ is the trainable parameters of encoders $f_{\textrm{Enc}}^r, f_{\textrm{Enc}}^a$ and the cross-mapping layers $f_{\textrm{AR}}, f_{\textrm{RA}}$, $\theta_{\textrm{Dis}}$ collects parameters of all $T$ pairs of discriminators $D_a^k, D_r^k$. The overall objective of the multiomics alignment $f$ is
\begin{equation}
    \mathcal{L}_{\textrm{Int}} = \mathcal{L}_{\textrm{Rec}} + \gamma \mathcal{L}_{\textrm{Adv}}
\end{equation}
where $\gamma$ is a hyperparameter to weigh the adversarial loss. After the training, the multiomics alignment module $f$ is defined as
\begin{equation}
    f(\mathbf{x};\theta_f) = f_{\textrm{Dec}}^r(f_{\textrm{AR}}(f_{\textrm{Enc}}^a(\mathbf{W}^a_{\textrm{Emb}}(\mathbf{x})))).
\end{equation}

Once training is complete, the multiomics alignment module $f$ can be used to transform ATAC-seq data into the corresponding RNA-seq format.

\paragraph{Performance of Multiomics Alignment}
We assessed the model's accuracy on a single-cell multiomics dataset. We selected marker genes from a previous study \cite{Lake2016} and compared the mean expressions between cell types and between the observed and translated cohorts (Figure \ref{fig:fig3}A). Marker genes, which are highly indicative of each cell type, served as category labels in this evaluation. Focusing on key marker genes, the UMAP showed consistent findings, with the translated expression highlighting these cell types and high correlations between observed and translated data (Figure \ref{fig:fig3}B). 

\begin{figure*}[t]
    \centering
    \begin{subfigure}[b]{0.4\textwidth}
         \centering
         \includegraphics[width=\textwidth]{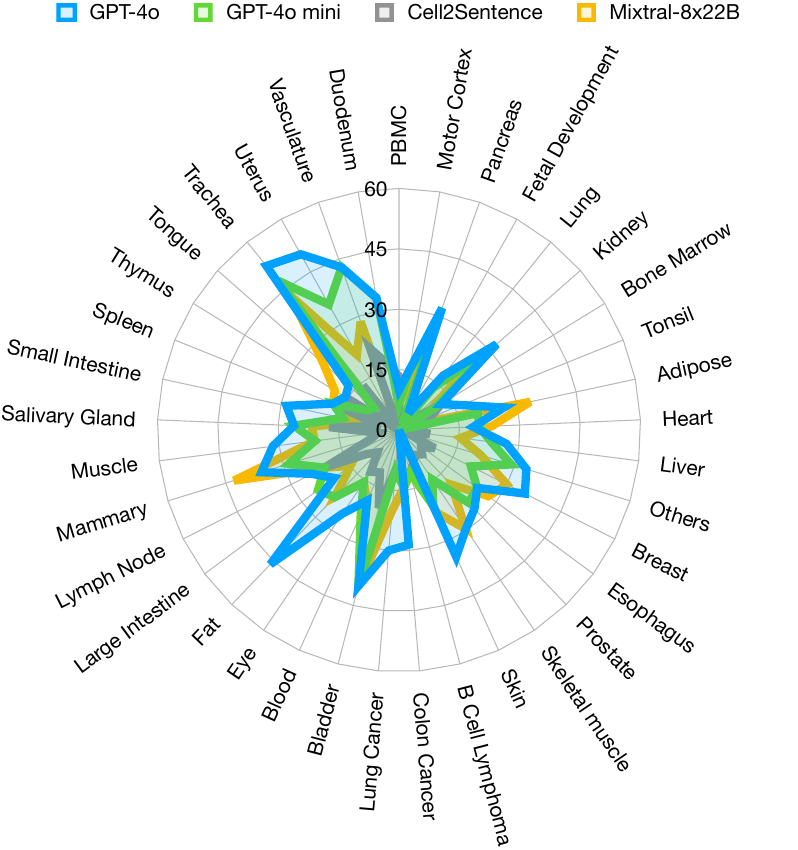}
         % \caption{Zero-Shot Prompting Strategy}
         % \label{fig:y equals x}
     \end{subfigure}
     % \hfill
     % \\
    \begin{subfigure}[b]{0.4\textwidth}
         \centering
         \includegraphics[width=\textwidth]{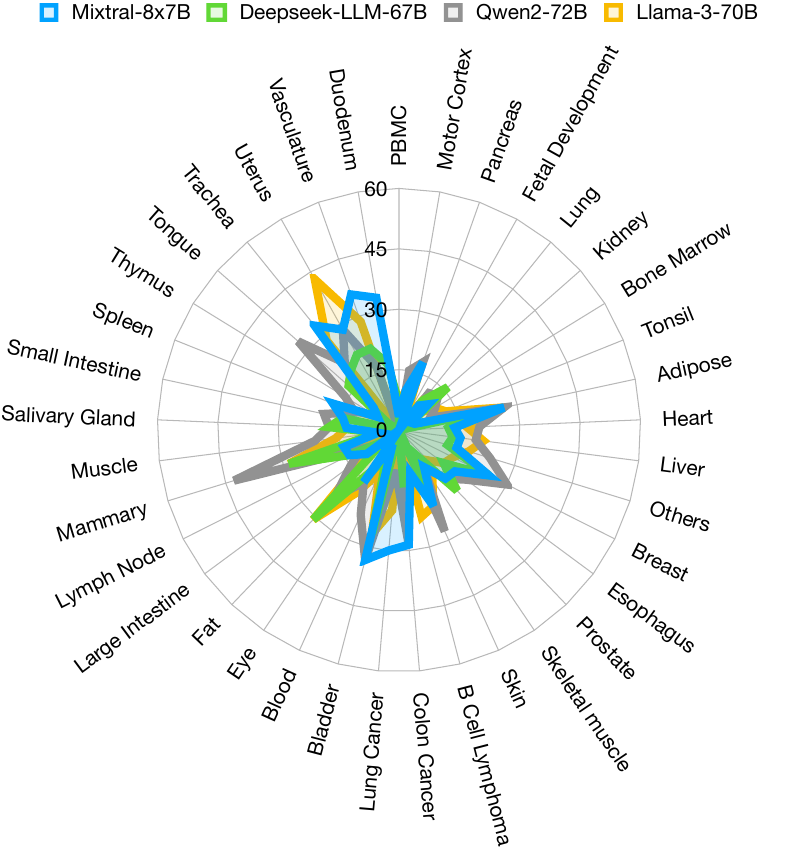}
         % \caption{Zero-Shot Chain-of-Thought Prompting Strategy}
         % \label{fig:y equals x}
     \end{subfigure}
\caption{The EM evaluation result per tissue of the \benchmarkName{} benchmark.}
    \label{fig:nature_method_em_per_tissue}
\end{figure*}

\begin{figure*}[!htbp]
    \centering
    \begin{subfigure}[b]{0.4\textwidth}
         \centering
         \includegraphics[width=\textwidth]{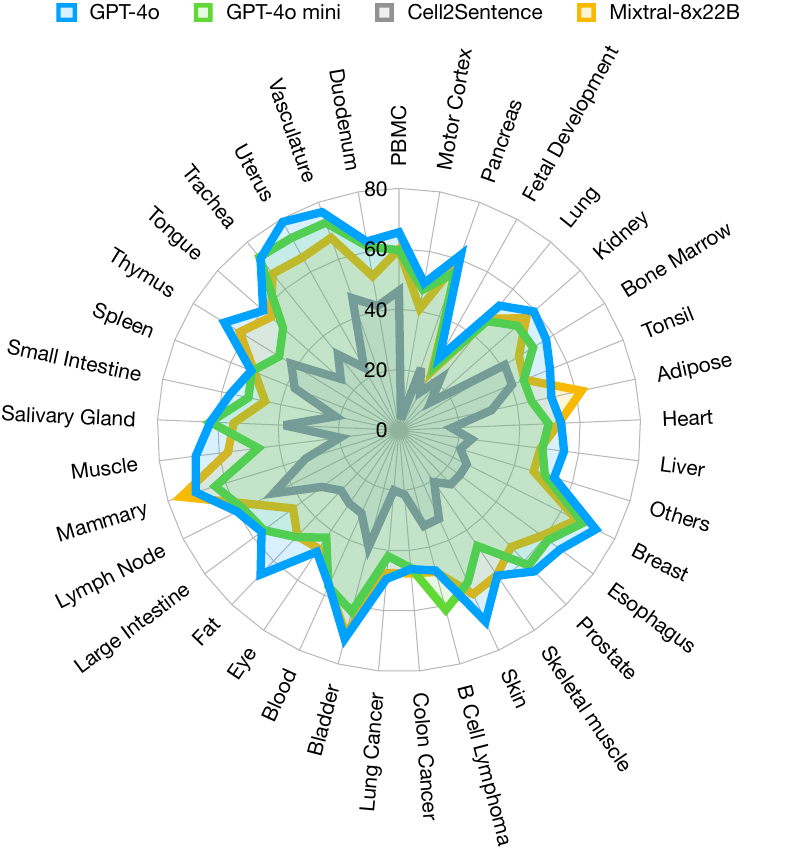}
         % \caption{Zero-Shot Prompting Strategy}
         % \label{fig:y equals x}
     \end{subfigure}
     % \hfill
     % \\
    \begin{subfigure}[b]{0.4\textwidth}
         \centering
         \includegraphics[width=\textwidth]{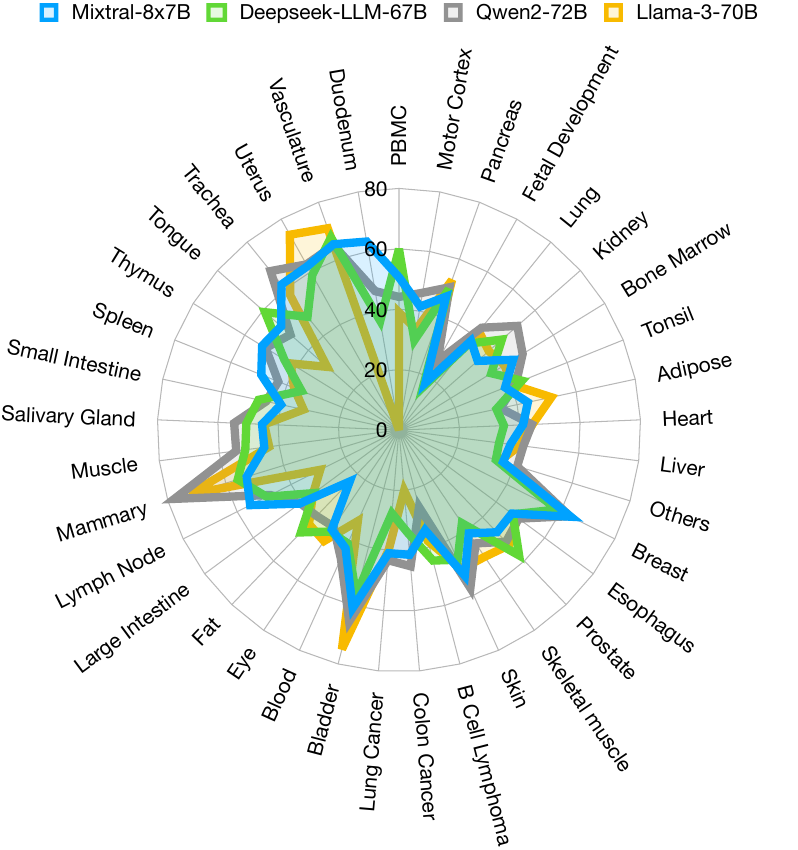}
         % \caption{Zero-Shot Chain-of-Thought Prompting Strategy}
         % \label{fig:y equals x}
     \end{subfigure}
\caption{The F1 evaluation result per tissue of the \benchmarkName{} benchmark.}
    \label{fig:nature_method_f1_per_tissue}
\end{figure*}

\section{Implementation Details}
\label{app:implementation_details}
For our experiments, we used a consistent set of hyperparameters across all instruction-tuned models, adhering to commonly accepted values from previous studies~\cite{wolf2019huggingface}. Specifically, we set the temperature at 0.6, top\_p at 0.9, and top\_k at 50. 
The temperature parameter adjusts the randomness during sampling, top\_p filters out tokens with lower probabilities, and top\_k limits the sampling process to the $k$ most probable tokens. 
The detailed method for cleansing and extracting the annotation result from the response is described in Appendix~\ref{app:answer_cleansing}. For Cell2Sentence, we followed the instruction finetuning method proposed in~\cite{levinecell2sentence} to train the pretrained Cell2Sentence model using the instruction-following dataset provided by the authors. Our code is available in the Supplementary Materials.

\section{Detailed Evaluation Results}
\label{app:results_on_tissues}

\paragraph{Evaluation Result per Tissue} The detailed EM/F1 evaluation result per tissue on the \benchmarkName{} benchmark are presented in Figure~\ref{fig:nature_method_em_per_tissue} and~\ref{fig:nature_method_f1_per_tissue} respectively.

\begin{figure*}[!htbp]
    \centering
    \begin{subfigure}[b]{0.4\textwidth}
         \centering
         \includegraphics[width=\textwidth]{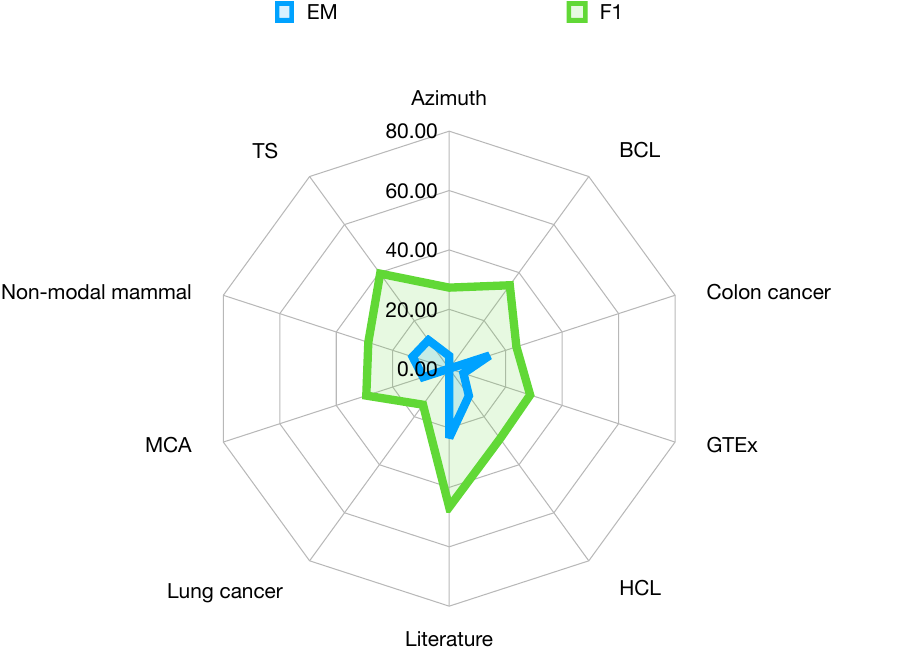}
         \caption{DeepSeek-LLM-67B}
         % \label{fig:y equals x}
     \end{subfigure}
    \begin{subfigure}[b]{0.4\textwidth}
         \centering
         \includegraphics[width=\textwidth]{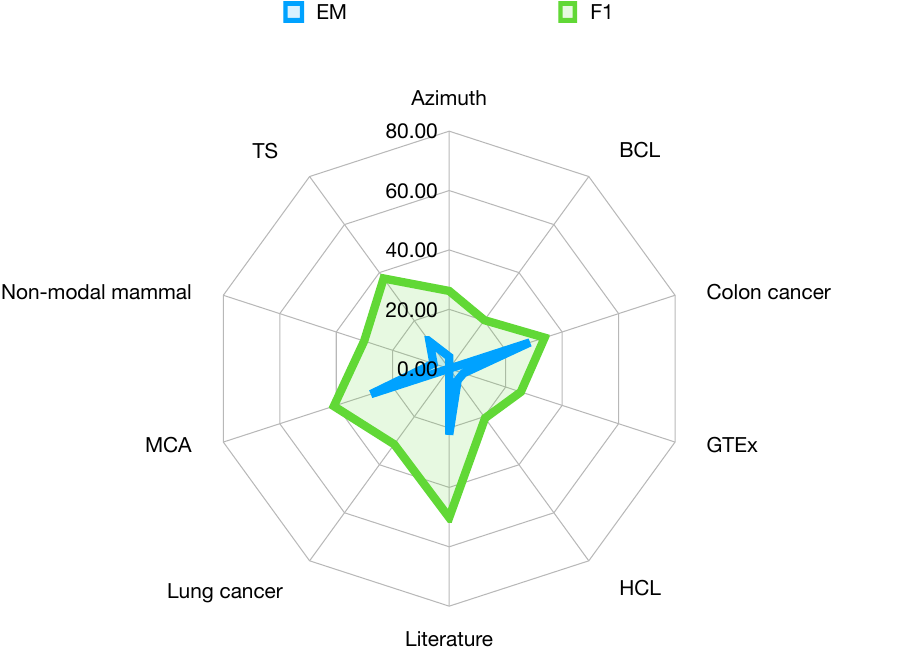}
         \caption{Qwen2-72B}
         % \label{fig:y equals x}
     \end{subfigure}\\
     \begin{subfigure}[b]{0.4\textwidth}
         \centering
         \includegraphics[width=\textwidth]{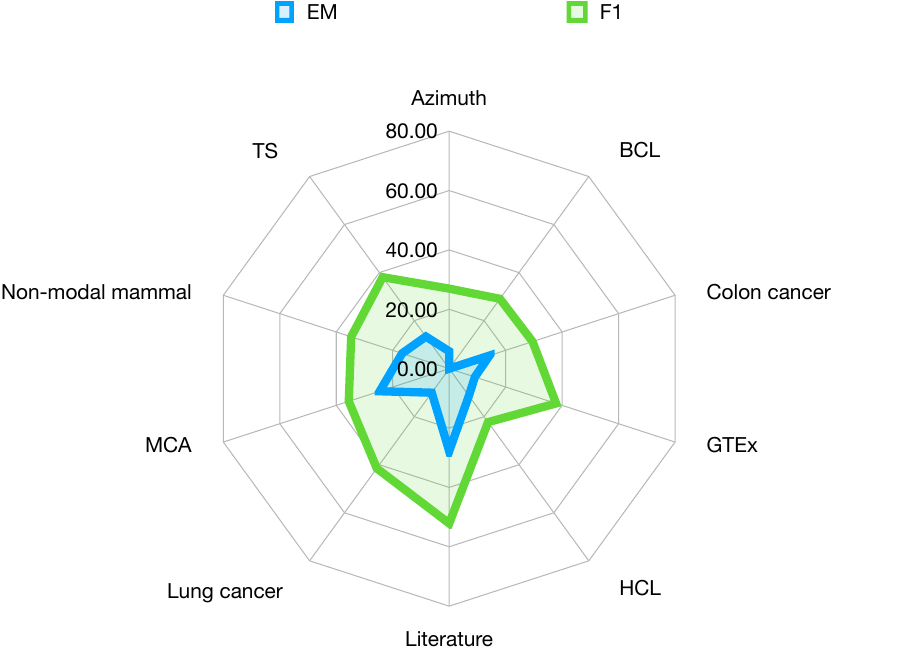}
         \caption{Llama-3-70B}
         % \label{fig:y equals x}
     \end{subfigure}
    \begin{subfigure}[b]{0.4\textwidth}
         \centering
         \includegraphics[width=\textwidth]{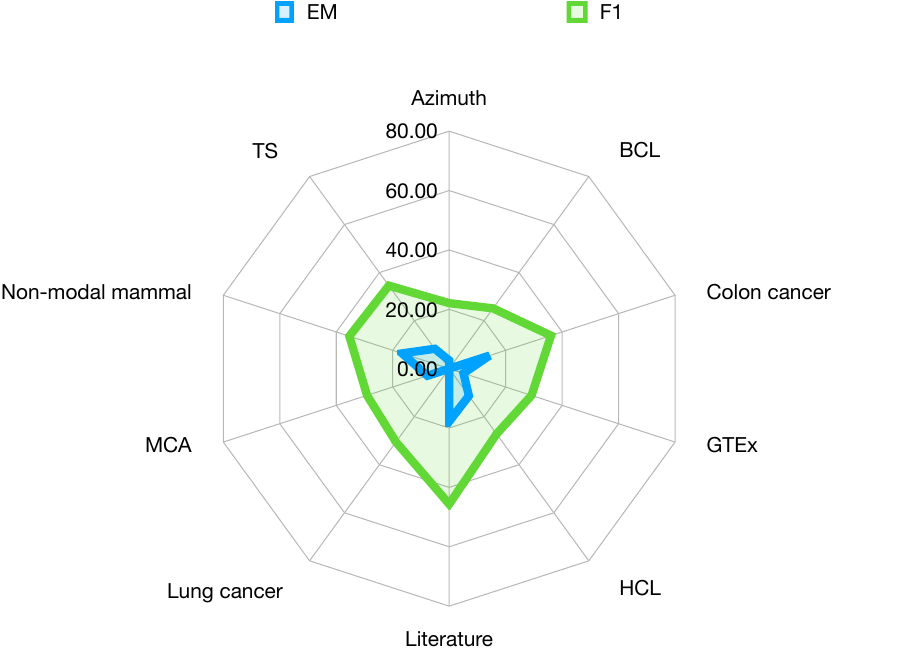}
         \caption{Mixtral-8×7B}
         % \label{fig:y equals x}
     \end{subfigure}
     \begin{subfigure}[b]{0.4\textwidth}
         \centering
         \includegraphics[width=\textwidth]{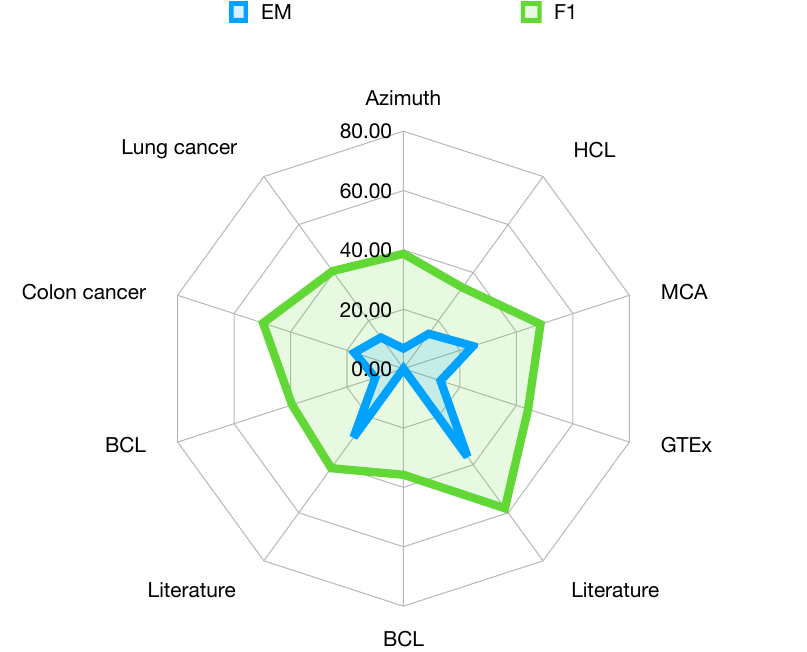}
         \caption{Mixtral-8×22B}
         % \label{fig:y equals x}
     \end{subfigure}
    \begin{subfigure}[b]{0.4\textwidth}
         \centering
         \includegraphics[width=\textwidth]{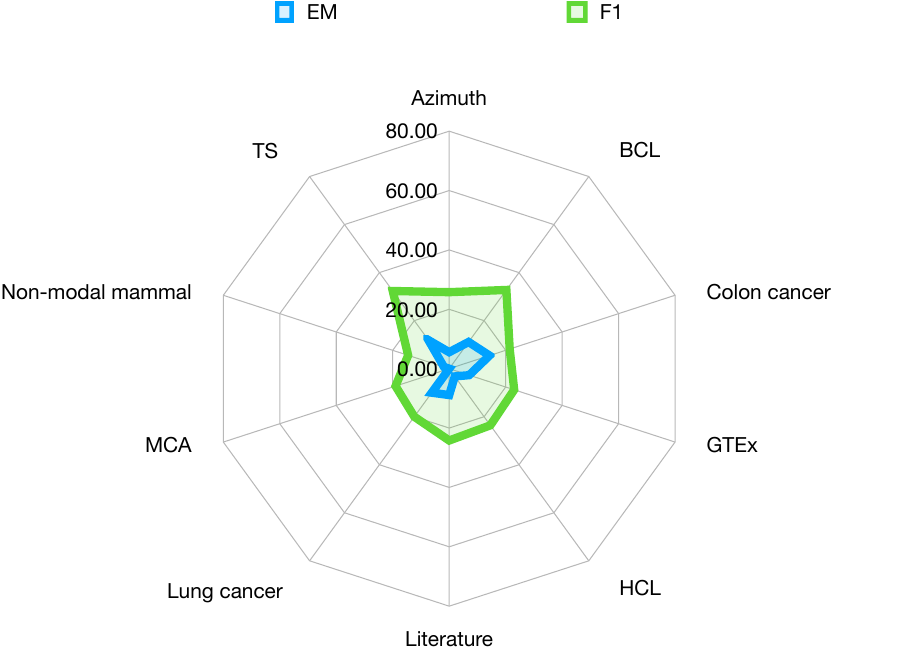}
         \caption{Cell2Sentence}
         % \label{fig:y equals x}
     \end{subfigure}
     \begin{subfigure}[b]{0.4\textwidth}
         \centering
         \includegraphics[width=\textwidth]{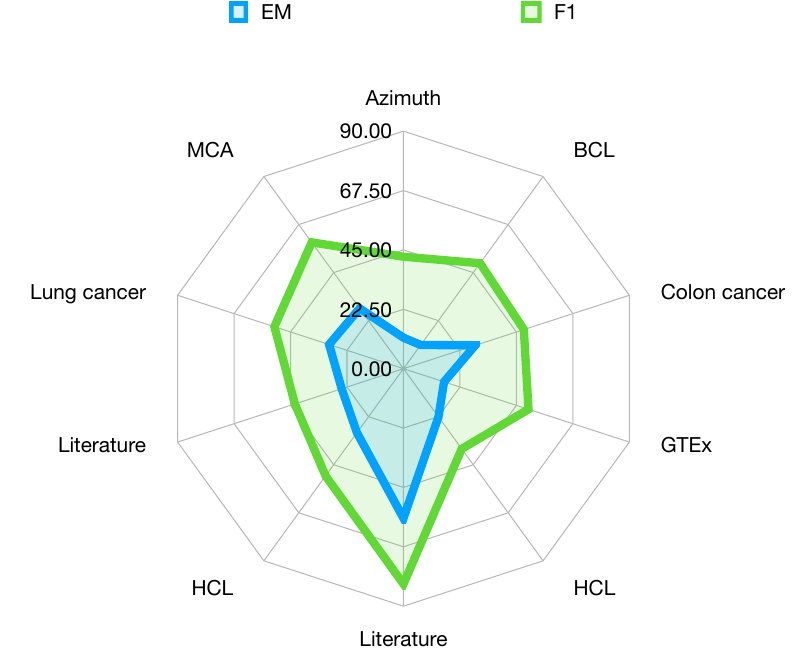}
         \caption{GPT-4o mini}
         % \label{fig:y equals x}
     \end{subfigure}
    \begin{subfigure}[b]{0.4\textwidth}
         \centering
         \includegraphics[width=\textwidth]{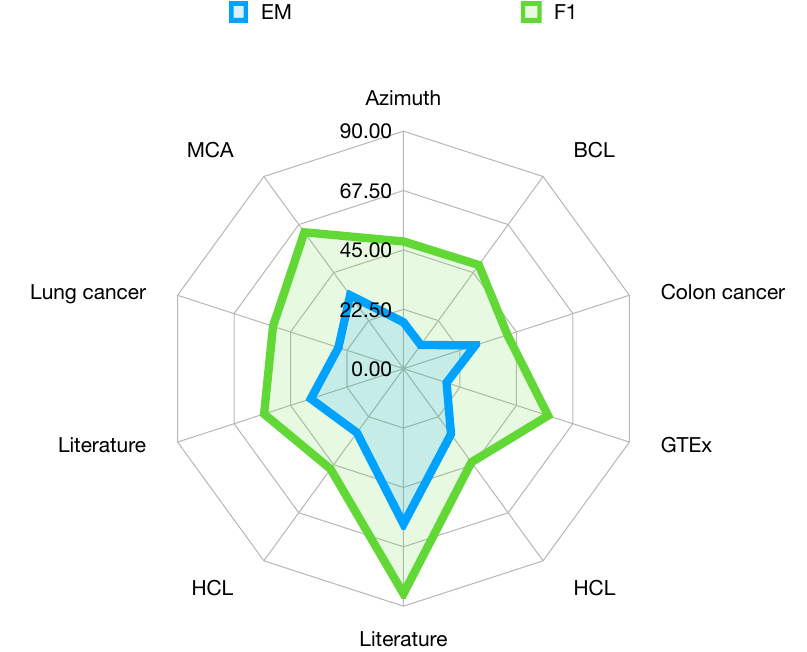}
         \caption{GPT-4o}
         % \label{fig:y equals x}
     \end{subfigure}
\caption{The EM and F1 evaluation result per dataset of the \benchmarkName{} benchmark using the zero-shot prompt.}
    \label{fig:nature_method_squad_per_dataset}
\end{figure*}

\begin{figure*}[!htbp]
    \centering
    \begin{subfigure}[b]{0.4\textwidth}
         \centering
         \includegraphics[width=\textwidth]{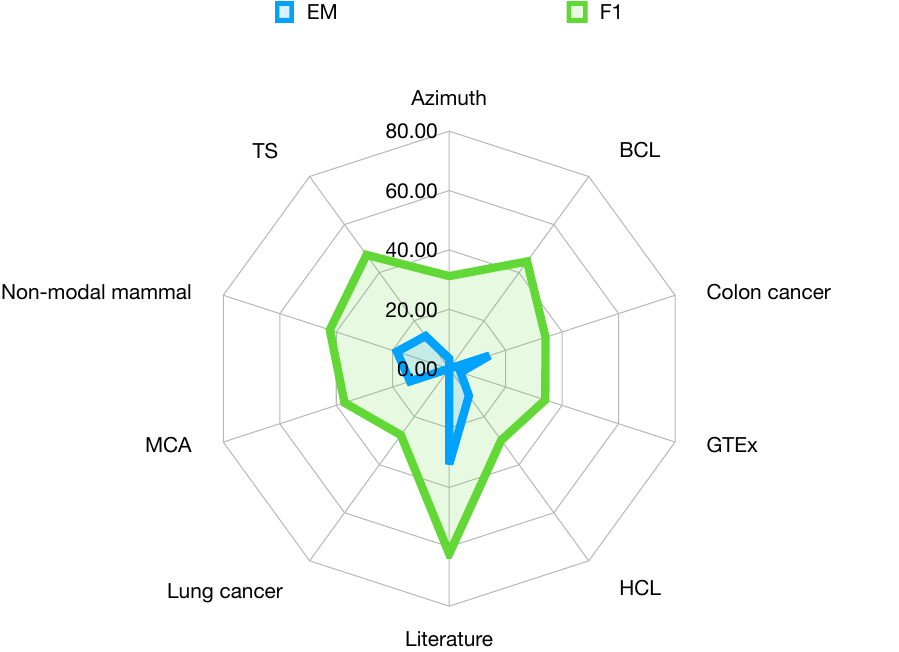}
         \caption{DeepSeek-LLM-67B}
         % \label{fig:y equals x}
     \end{subfigure}
    \begin{subfigure}[b]{0.4\textwidth}
         \centering
         \includegraphics[width=\textwidth]{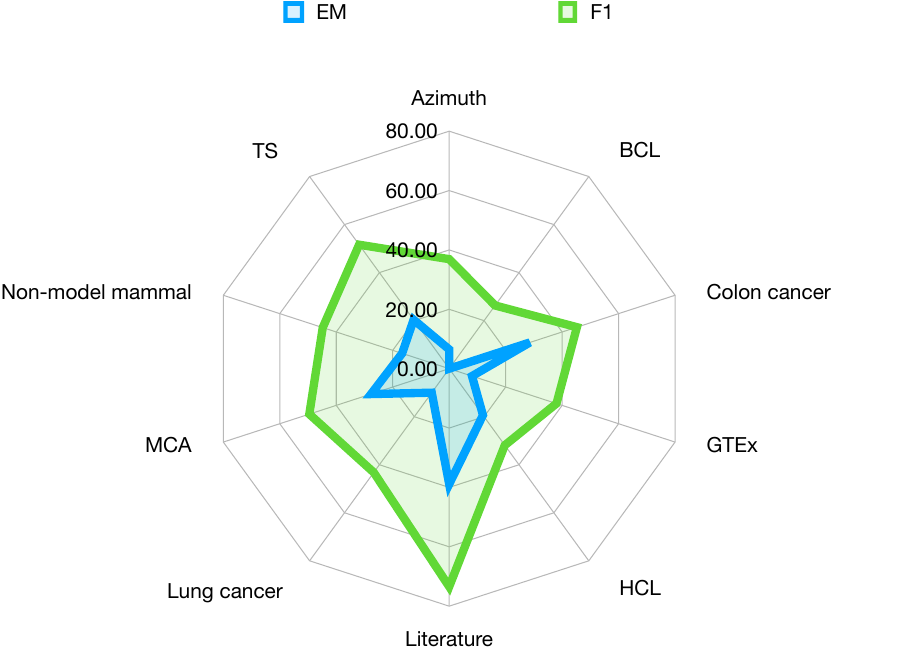}
         \caption{Qwen2-72B}
         % \label{fig:y equals x}
     \end{subfigure}\\
     \begin{subfigure}[b]{0.4\textwidth}
         \centering
         \includegraphics[width=\textwidth]{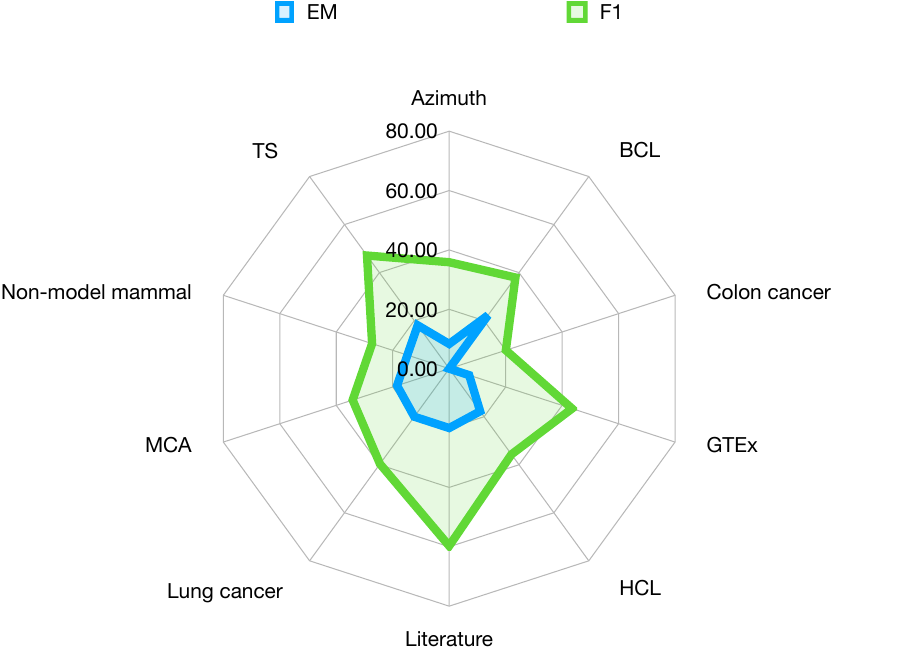}
         \caption{Llama-3-70B}
         % \label{fig:y equals x}
     \end{subfigure}
    \begin{subfigure}[b]{0.4\textwidth}
         \centering
         \includegraphics[width=\textwidth]{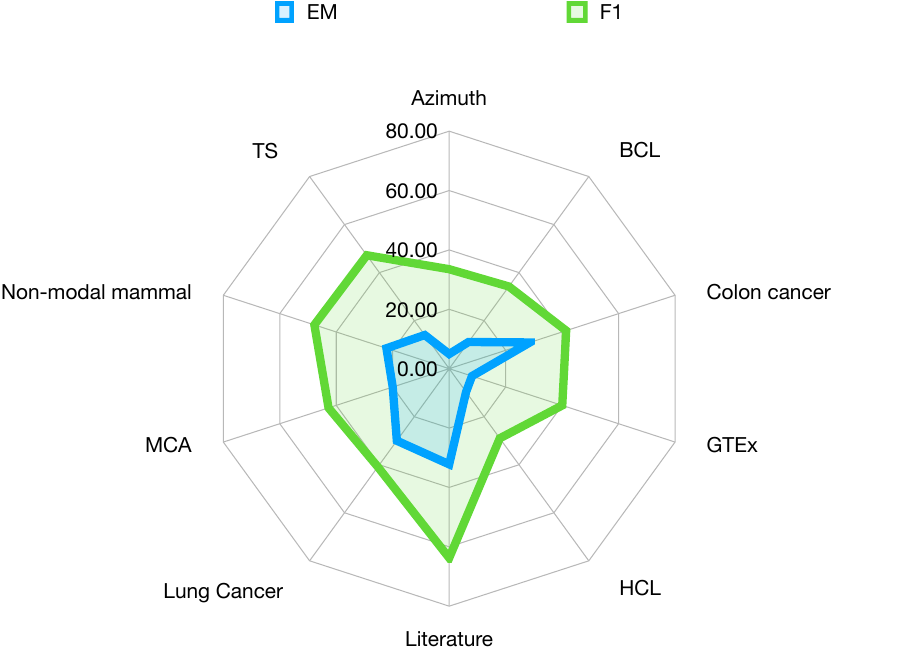}
         \caption{Mixtral-8×7B}
         % \label{fig:y equals x}
     \end{subfigure}
     \begin{subfigure}[b]{0.4\textwidth}
         \centering
         \includegraphics[width=\textwidth]{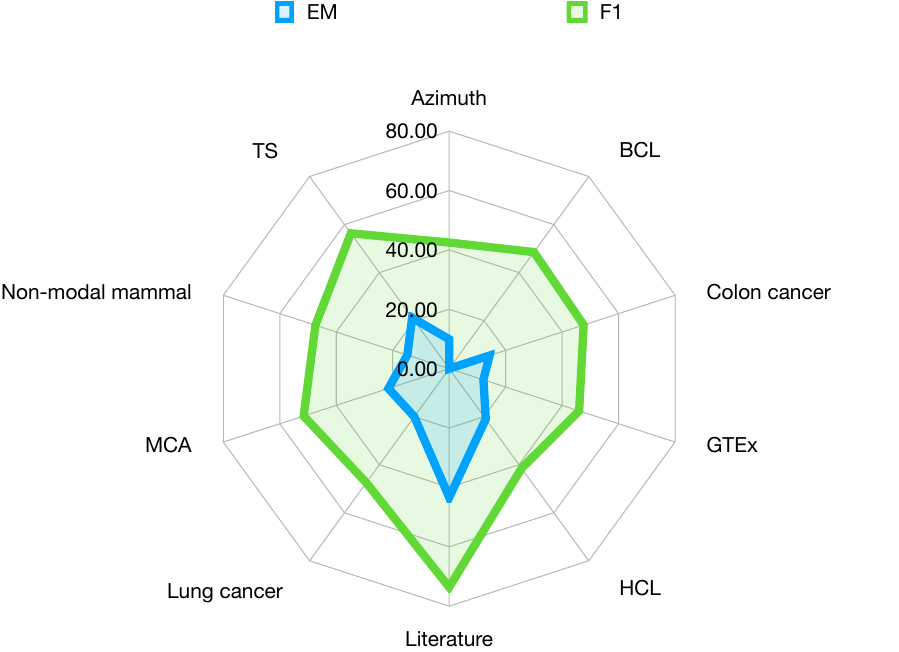}
         \caption{Mixtral-8×22B}
         % \label{fig:y equals x}
     \end{subfigure}
    \begin{subfigure}[b]{0.4\textwidth}
         \centering
         \includegraphics[width=\textwidth]{latex/figure/nature_method_per_dataset/cell2sent-cot.pdf}
         \caption{Cell2Sentence}
         % \label{fig:y equals x}
     \end{subfigure}
     \begin{subfigure}[b]{0.4\textwidth}
         \centering
         \includegraphics[width=\textwidth]{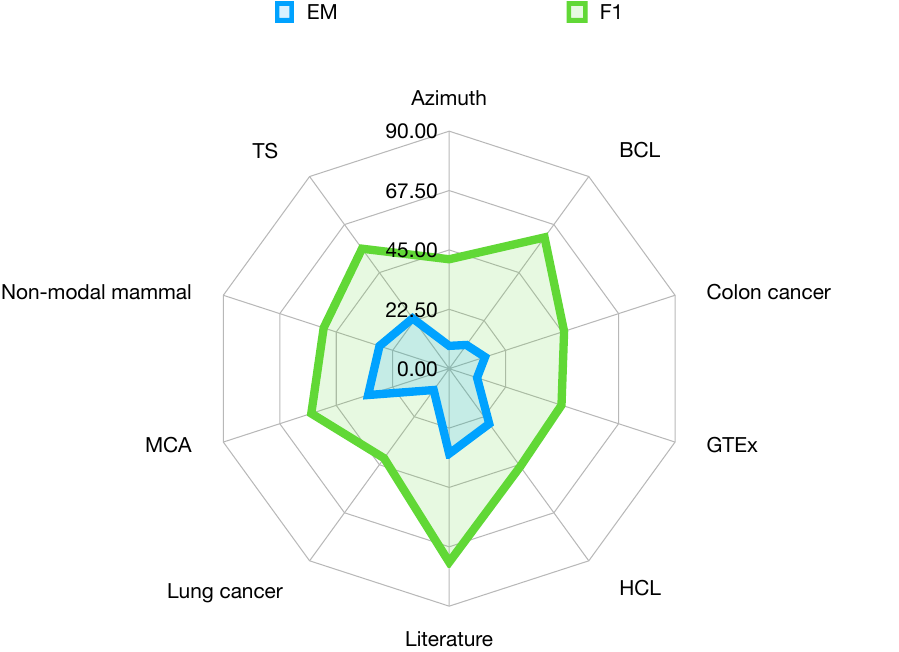}
         \caption{GPT-4o mini}
         % \label{fig:y equals x}
     \end{subfigure}
    \begin{subfigure}[b]{0.4\textwidth}
         \centering
         \includegraphics[width=\textwidth]{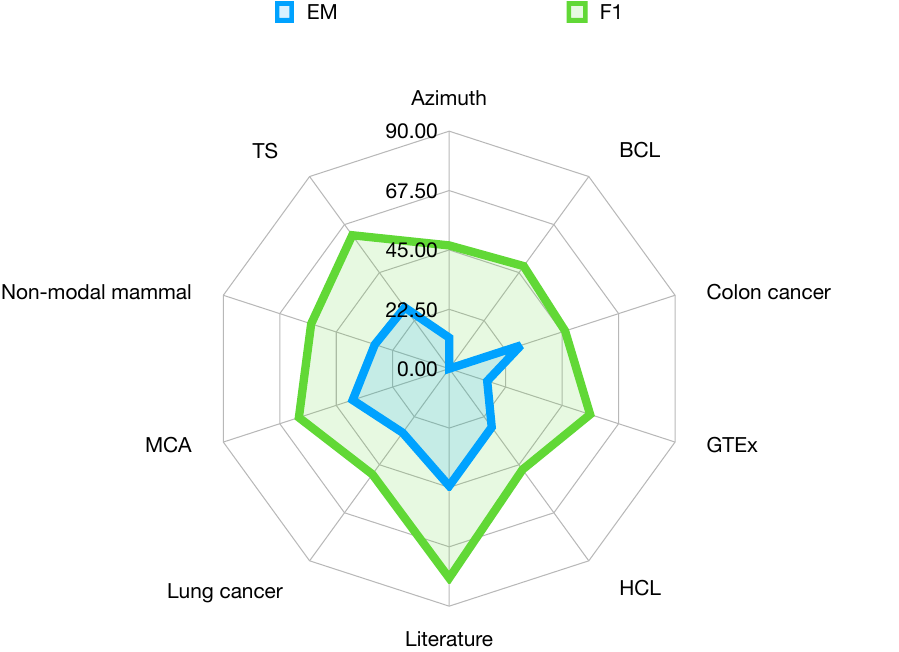}
         \caption{GPT-4o}
         % \label{fig:y equals x}
     \end{subfigure}
\caption{The EM and F1 evaluation result per dataset of the \benchmarkName{} benchmark using the zero-shot CoT prompt.}
    \label{fig:nature_method_squad_per_dataset_cot}
\end{figure*}

\paragraph{Evaluation Result per Dataset}
\label{app:results_on_datasets}
The EM/F1 evaluation results per dataset on \benchmarkName{} using the zero-shot and zero-shot CoT prompt are in Figure~\ref{fig:nature_method_squad_per_dataset} and~\ref{fig:nature_method_squad_per_dataset_cot} respectively.

\section{Cell Type Annotation Example}
\label{app:annotation}
The cell type annotation processes performed by LLMs are detailed in Table~\ref{tab:chat1} to~\ref{tab:chat5}. LLMs can analyze each marker provided in the list using the zero-shot chain-of-thought prompt and summarize the final cell type. For brevity, the system message \textit{``You are a biology expert who always responds the cell type annotation result by carefully considering the markers provided by the user.''} is omitted.

\begin{table*}[!htbp]
\small
\centering
\begin{tabular}{p{\textwidth}}
\toprule
\textbf{User} \\
Given the following markers \textcolor{blue}{[MS4A1, TNFRSF13B, IGHM, IGHD, AIM2, CD79A, LINC01857, RALGPS2, BANK1, CD79B]}, what is the cell type in \textcolor{blue}{PBMC} corresponding to these markers? Let's think step by step.
\\ \midrule
\textbf{Assistant}            \\
To determine the cell type in peripheral blood mononuclear cells (PBMC) corresponding to the given markers, we need to analyze each marker and its known associations with specific cell types. \\\\

Here are the markers provided:\\
1. MS4A1 (CD20): This is a well-known marker for B cells. \\
2. TNFRSF13B (TACI): This is a receptor involved in B cell development and function. \\
3. IGHM: This gene encodes the constant region of the mu heavy chain of immunoglobulin M (IgM), which is expressed on immature and mature B cells. \\
4. IGHD: This gene encodes the delta heavy chain of immunoglobulin D (IgD), which is expressed on mature B cells. \\
5. AIM2: This gene is involved in the immune response and can be expressed in various cell types, including B cells. \\
6. CD79A: This gene encodes a component of the B cell receptor complex, essential for B cell development and signaling. \\
7. LINC01857: This is a long non-coding RNA, and its specific association with cell types is less clear, but it can be expressed in B cells. \\
8. RALGPS2: This gene is involved in cell signaling and can be expressed in various cell types, including B cells. \\
9. BANK1: This gene is specifically expressed in B cells and is involved in B cell receptor signaling. \\
10. CD79B: This gene encodes another component of the B cell receptor complex, essential for B cell development and signaling.  \\\\

Step-by-Step Analysis:  \\
1. MS4A1 (CD20): Strongly indicates B cells. \\
2. TNFRSF13B (TACI): Associated with B cells. \\
3. IGHM: Indicates B cells, particularly those expressing IgM. \\
4. IGHD: Indicates mature B cells expressing IgD. \\
5. AIM2: Can be expressed in B cells, though not exclusive. \\
6. CD79A: Essential for B cell receptor function, indicating B cells. \\
7. LINC01857: Less specific, but can be expressed in B cells. \\
8. RALGPS2: Can be expressed in B cells. \\
9. BANK1: Specifically expressed in B cells. \\
10. CD79B: Essential for B cell receptor function, indicating B cells.\\\\

Conclusion: The combination of these markers strongly suggests that the cell type in PBMC corresponding to these markers is B cells. The presence of markers like MS4A1 (CD20), CD79A, CD79B, IGHM, and IGHD, which are all critical for B cell identification and function, supports this conclusion.
\\ \midrule
\textbf{User}            \\
In summary, the most likely cell type (directly return one cell type name) is
                \\ \midrule
\textbf{Assistant}            \\
B cells~~\cmark
            \\ \bottomrule
\end{tabular}
\caption{Example of LLM annotating cell type using the zero-shot chain-of-thought prompting.}
\label{tab:chat1}
\end{table*}

\begin{table*}[!htbp]
\small
\centering
\begin{tabular}{p{\textwidth}}
\toprule
\textbf{User} \\
Given the following markers \textcolor{blue}{[CLEC9A, PPY, AL118508.1, LINC02206, LINC01724, MIR1273H, CEACAM3, CLCN1, AC092809.2, CYP2E1]}, what is the cell type in \textcolor{blue}{Thymus} corresponding to these markers? Let's think step by step.
\\ \midrule
\textbf{Assistant}            \\
To determine the cell type in the thymus corresponding to the given markers, we need to analyze each marker and its known associations. \\\\Here are the steps to identify the cell type:\\1. CLEC9A: This marker is typically associated with dendritic cells, particularly a subset known as CD141+ (BDCA-3+) dendritic cells. These cells are involved in antigen presentation and are known for their role in the immune response.\\2. PPY (Pancreatic Polypeptide): This marker is generally associated with pancreatic cells, specifically PP cells in the pancreas. However, its presence in the thymus might be less common and could indicate a specific subset of cells or a particular state of differentiation.\\3. AL118508.1, LINC02206, LINC01724, MIR1273H, AC092809.2: These are non-coding RNAs (long intergenic non-coding RNAs and microRNAs). The specific expression patterns of these non-coding RNAs can be cell-type specific, but they are less commonly used as primary markers for identifying cell types without additional context.\\4. CEACAM3: This marker is typically associated with granulocytes, particularly neutrophils. It is involved in the immune response and pathogen recognition.\\5. CLCN1: This gene encodes a chloride channel protein, which is primarily associated with muscle cells. Its relevance in the thymus might be less direct.\\6. CYP2E1: This gene encodes an enzyme involved in drug metabolism and is primarily expressed in the liver. Its expression in the thymus could be indicative of specific metabolic activities but is not a primary marker for thymic cell types.\\Given the context of the thymus, the most relevant marker here is CLEC9A, which points towards dendritic cells. The presence of CEACAM3 might suggest some involvement of granulocytes or neutrophils, but this is less typical for the thymus.\\\\Considering the primary role of the thymus in T-cell development and the presence of dendritic cells for antigen presentation, the most likely cell type corresponding to these markers in the thymus is a subset of dendritic cells.\\Therefore, the cell type in the thymus corresponding to these markers is likely dendritic cells, specifically the CD141+ (BDCA-3+) subset.
\\ \midrule
\textbf{User}            \\
In summary, the most likely cell type (directly return one cell type name) is
                \\ \midrule
\textbf{Assistant}            \\
Dendritic cells~~\cmark~~(Correct Answer: Myeloid dendritic cell)
            \\ \bottomrule
\end{tabular}
\caption{Example of LLM annotating cell type using the zero-shot chain-of-thought prompting.}
\label{tab:chat2}
\end{table*}

\begin{table*}[!htbp]
\small
\centering
\begin{tabular}{p{\textwidth}}
\toprule
\textbf{User} \\
Given the following markers \textcolor{blue}{[Cftr, Epcam, Onecut1, Tm4sf4]}, what is the cell type in \textcolor{blue}{Liver} corresponding to these markers? Let's think step by step.
\\ \midrule
\textbf{Assistant}            \\
To determine the cell type in the liver corresponding to the given markers [Cftr, Epcam, Onecut1, Tm4sf4], we need to analyze each marker and its known associations with specific cell types in the liver.\\1. Cftr (Cystic Fibrosis Transmembrane Conductance Regulator):- Cftr is a protein that functions as a channel for chloride ions across epithelial cell membranes. It is known to be expressed in various epithelial cells, including those in the liver.\\2. Epcam (Epithelial Cell Adhesion Molecule): - Epcam is a marker for epithelial cells. It is commonly used to identify epithelial cells in various tissues, including the liver.\\3. Onecut1 (One Cut Homeobox 1): - Onecut1, also known as HNF-6 (Hepatocyte Nuclear Factor 6), is a transcription factor that plays a crucial role in the development and function of liver cells, particularly hepatocytes and cholangiocytes (bile duct cells).\\4. Tm4sf4 (Transmembrane 4 L Six Family Member 4):  - Tm4sf4 is a marker that is often associated with epithelial cells, including those in the liver. It has been implicated in the regulation of cell proliferation and differentiation.\\\\Given the combination of these markers:\\- Cftr and Epcam suggest an epithelial origin.\\- Onecut1 (HNF-6) is indicative of liver-specific epithelial cells, particularly hepatocytes and cholangiocytes.\\- Tm4sf4 further supports the epithelial nature of the cells.\\\\Considering the liver context and the expression of these markers, the most likely cell type corresponding to these markers is cholangiocytes (bile duct epithelial cells). Cholangiocytes are known to express Cftr, Epcam, Onecut1, and Tm4sf4, making them the best match for the provided markers.\\\\Therefore, the cell type in the liver corresponding to the markers [Cftr, Epcam, Onecut1, Tm4sf4] is cholangiocytes.
\\ \midrule
\textbf{User}            \\
In summary, the most likely cell type (directly return one cell type name) is
                \\ \midrule
\textbf{Assistant}            \\
Cholangiocytes~~\cmark
            \\ \bottomrule
\end{tabular}
\caption{Example of LLM annotating cell type using the zero-shot chain-of-thought prompting.}
\label{tab:chat3}
\end{table*}

\begin{table*}[!htbp]
\small
\centering
\begin{tabular}{p{\textwidth}}
\toprule
\textbf{User} \\
Given the following markers \textcolor{blue}{[Lrp2, Pdzk1, Slco3a1]}, what is the cell type in \textcolor{blue}{Kidney} corresponding to these markers? Let's think step by step.
\\ \midrule
\textbf{Assistant}            \\
To determine the cell type in the kidney corresponding to the markers Lrp2, Pdzk1, and Slco3a1, we need to analyze each marker and their known associations with specific cell types in the kidney.\\1. Lrp2 (Low-density lipoprotein receptor-related protein 2):- Lrp2, also known as megalin, is a large endocytic receptor that is highly expressed in the proximal tubule cells of the kidney. It plays a crucial role in the reabsorption of various molecules from the filtrate in the proximal tubule.\\

2. Pdzk1 (PDZ domain containing 1): - Pdzk1 is a scaffolding protein that interacts with several membrane proteins and is also highly expressed in the proximal tubule cells of the kidney. It is involved in the regulation of various transporters and channels in these cells.\\

3. Slco3a1 (Solute carrier organic anion transporter family member 3A1):  - Slco3a1 is a member of the organic anion transporting polypeptides (OATPs) family, which are involved in the transport of a wide range of endogenous and exogenous organic anions. This transporter is also expressed in the proximal tubule cells of the kidney.\\\\

Given the expression patterns and functions of these markers, we can conclude that:\\- Lrp2 is a marker for proximal tubule cells.\\- Pdzk1 is a marker for proximal tubule cells.\\- Slco3a1 is a marker for proximal tubule cells.
\\\\
Conclusion:\\The cell type in the kidney corresponding to the markers Lrp2, Pdzk1, and Slco3a1 is the proximal tubule cell.
\\ \midrule
\textbf{User}            \\
In summary, the most likely cell type (directly return one cell type name) is
                \\ \midrule
\textbf{Assistant}            \\
Proximal tubule cell~~\cmark
            \\ \bottomrule
\end{tabular}
\caption{Example of LLM annotating cell type using the zero-shot chain-of-thought prompting.}
\label{tab:chat4}
\end{table*}

\begin{table*}[!htbp]
\small
\centering
\begin{tabular}{p{\textwidth}}
\toprule
\textbf{User} \\
Given the following markers \textcolor{blue}{[RF00322-9, MIR624, FO624990.1, RNU7-134P, RF00019-186, ZNF587P1, MAL2, PINX1, YWHAQP7, AC112907.2]}, what is the cell type in \textcolor{blue}{Bone Marrow} corresponding to these markers? Let's think step by step.
\\ \midrule
\textbf{Assistant}            \\
To determine the cell type in Bone Marrow corresponding to the given markers, we need to analyze each marker and its known associations with specific cell types. Let's go through each marker step by step:\\1. RF00322-9: This is a non-coding RNA (ncRNA) identifier. Non-coding RNAs can be involved in various regulatory functions but are not specific to a particular cell type without additional context.\\2. MIR624: This is a microRNA (miRNA). miRNAs are small non-coding RNAs that regulate gene expression. MIR624 has been implicated in various cellular processes, including differentiation and proliferation, but it is not specific to a particular cell type in the bone marrow.\\3. FO624990.1: This appears to be a genomic locus identifier, which does not provide direct information about cell type without further context.\\4. RNU7-134P: This is a small nuclear RNA (snRNA) pseudogene. Pseudogenes are typically non-functional remnants of genes and do not provide specific information about cell type.\\5. RF00019-186: This is another non-coding RNA identifier, similar to RF00322-9, and does not provide specific cell type information without additional context.\\6. ZNF587P1: This is a zinc finger protein pseudogene. Zinc finger proteins are involved in DNA binding and transcriptional regulation, but pseudogenes are generally non-functional and do not indicate a specific cell type.\\7. MAL2: This gene encodes a protein involved in the transport of proteins to the apical membrane in polarized epithelial cells. While MAL2 is not specific to bone marrow, it has been associated with certain hematopoietic cells.\\8. PINX1: This gene encodes a protein that interacts with telomerase and is involved in telomere maintenance. It is expressed in various cell types, including hematopoietic cells.\\9. YWHAQP7: This is a pseudogene related to the 14-3-3 protein family, which is involved in signal transduction. Pseudogenes do not provide specific cell type information.\\10. AC112907.2: This is a long non-coding RNA (lncRNA) identifier. lncRNAs can have various regulatory roles but are not specific to a particular cell type without additional context.\\\\
Given the markers provided, most of them are non-coding RNAs, pseudogenes, or general regulatory proteins that do not point to a specific cell type in the bone marrow. However, the presence of MAL2 and PINX1 suggests a potential association with hematopoietic cells.\\Considering the context of bone marrow and the markers provided, the most likely cell type could be a hematopoietic progenitor cell or a hematopoietic stem cell. These cells are characterized by their ability to differentiate into various blood cell lineages and are known to express a variety of regulatory RNAs and proteins involved in cell differentiation and proliferation.\\\\
Therefore, based on the markers provided, the cell type in Bone Marrow corresponding to these markers is likely a hematopoietic progenitor cell or hematopoietic stem cell.
\\ \midrule
\textbf{User}            \\
In summary, the most likely cell type (directly return one cell type name) is
                \\ \midrule
\textbf{Assistant}            \\
Hematopoietic progenitor cell~~\xmark~~(Correct Answer: Plasmablast)
            \\ \bottomrule
\end{tabular}
\caption{Example of LLM annotating cell type using the zero-shot chain-of-thought prompting.}
\label{tab:chat5}
\end{table*}
\end{document}